\documentclass[11pt]{article}

\usepackage[final]{acl}

\usepackage{times}
\usepackage{latexsym}

\usepackage[T1]{fontenc}

\usepackage[utf8]{inputenc}

\usepackage{microtype}

\usepackage{inconsolata}

\usepackage{graphicx}
\DeclareGraphicsExtensions{.pdf,.ai,.jpg,.png}
\setkeys{Gin}{pagebox=artbox}%

\usepackage{mathtools}

\usepackage{booktabs}
\usepackage{amssymb}
\usepackage{multirow}

\graphicspath{{emnlp26-attribute-anlaysis-and-steering-figures/}}

\newcommand{\bsfigure}[3][]{%
	\begin{figure}[t]
		\centering
		\includegraphics[#1]{#2}
		\caption{#3}\label{#2}%
 	 \end{figure}
}

\newcommand{\hwfigure}[3][t!]{%
	\begin{figure*}[#1]
		\centering
		\includegraphics[scale=1.0]{#2}
    		\caption{#3}\label{#2}%
  	\end{figure*}
}

\RequirePackage{type1cm}
\RequirePackage{color}
\RequirePackage{soul}
\setstcolor{blue}
\definecolor{violet}{rgb}{0.5,0.0,0.5}
\newsavebox\bscombox
\newcommand{\bscom}[3][]{%
	\sbox{\bscombox}{\fontsize{8}{9}\selectfont#1#2#3}
	\noindent
	\st{#2}{\selectfont
		\color{blue}#3\ifx\\#1\\\else{\fontsize{8}{9}\selectfont\color{violet}[#1]}\fi
	}
}

\definecolor{highlight1}{rgb}{0.95,0.95,0.95}
\definecolor{tgray}{rgb}{0.5,0.5,0.5}
\definecolor{tgray}{rgb}{0.5,0.5,0.5}



\begin{document}

\title{Attribute-Based Activation Steering of LLMs for \\ Group-Specific Explanation Generation}

\author{
	Leandra Fichtel \\
	Leibniz University Hannover \\
	\texttt{l.fichtel@ai.uni-hannover.de} \And
	Janek Prange \\
	Leibniz University Hannover \\
	\texttt{janek.prange@posteo.de} \AND
	Henning Wachsmuth \\
	Leibniz University Hannover, L3S Research Center \\
	\texttt{h.wachsmuth@ai.uni-hannover.de}
}

\maketitle
\begin{abstract}
To effectively enable people to understand new topics, explanations should be tailored to their backgrounds and abilities. Prompting alone has been shown to be insufficient for creating such explanations, and other computational methods are missing so far.
Therefore, this paper investigates whether LLMs can be steered to generate explanations that are tailored to~a specific group of people. To this end, we propose an approach that first identifies group-specific attributes in terms of explanatory style and knowledge of a specific target group. Building on activation engineering, it then computes \emph{attribute-based} steering vectors and adds them to the internal activations of an LLM during inference to enable a fine-grained steering. In our experiments, we assess the steering effectiveness in terms of specificity and factuality of the generated explanations. Additionally, we evaluate the explanations in a study with human experts from different target groups. Compared to prompting and state-of-the-art steering baselines, our approach tailors the explanations significantly better to the target group while~maintaining the best specificity-factuality balance.
\end{abstract}

\section{Introduction}
\label{sec:introduction}
The computational generation of natural language explanations has recently seen increased research interest, due to its importance for explainable AI \cite{schneider2019personalized, miller2019whathowwhy, lyu2024XAI, zytek2024XAI} and applications such as feedback-based writing support \cite{stahl:2024}. 
However, to ensure that explanations are effectively understood, the background and abilities of the explainee (i.e., the person being explained to) need to be taken into account. That is, explanations should ideally be personalized \cite{Sokol2020,Rong2024HumanCenteredXAI}. As explainee-specific characteristics are not always accessible, an alternative is to tailor the explanation to the explainee's target group \cite{schneider2019personalized}. For illustration, Figure~\ref{fig:sample} contrasts a generic explanation to an explanation tailored to game developers.

\bsfigure{example}{
Two exemplary explanations to the question ``How does DNA work?'': A \emph{generic explanation} generated by a prompt-based approach based on the question only, and a \emph{group-specific explanation} for game developers generated by the approach presented in this paper.\label{fig:sample}
}

While group-specific explanations could be generated by simply prompting large language models (LLMs) \cite{jeck25PersonalizedExpl, Zhang2025PersonalizedExplGen}, prompt-based steering is often highly sensitive to prompt formulation \cite{beck:2024} and achieves low steering effectiveness \cite{rooein2023knowYourAudience,siskou2025fingerpringtprompting}. For fine-tuning LLMs, group-specific explanation data is missing since most existing NLP research on human-written explanations lacks group-specific information about the explainees \cite{camburu2018eSNLI,rajani2019cosE, wang2020comVE,aggarwal2021ECQA}. Altogether, we are not aware of any computational method so far that generates high-quality explanations explicitly tailored to group-specific preferences.

In this paper, we tackle the task of group-specific explanation generation while avoiding the need for supervised explanation data. In particular, we study whether it is possible to computationally learn from explanations \emph{of} a specific target group how to explain a topic \emph{for} that group. Unlike text style transfer that targets the adjustment of style while preserving meaning \cite{jin:2022} and content transfer that does the opposite \cite{chen:2024}, the given problem requires modeling the background and abilities of the target group~\cite{wachsmuth2022wired}. Our hypothesis is that these are represented implicitly in the language that members of the group use and in what they talk about.
Extending the unsupervised method of \citet{patel2023}, we combine prompting and clustering to not only extract attributes related to explanatory style (e.g., \textit{The author includes code snippets}) but also to knowledge (e.g., \textit{The author understands game development}) from explanations written by group members (e.g., game developers).

The key idea of our approach is to generate explanations tailored to a specific target group by steering an LLM based on the most important attributes of that group. To enable this fine-grained steering, we employ activation engineering \cite{subramani:2022,turner2023actisteering,konen2024steeringstyle} by adding steering vectors to the LLM's internal activations for preselected layers during inference. Unlike existing approaches, we construct \emph{attribute-based} steering vectors from the activation vectors of the style and knowledge attributes to tailor the LLM's output specifically to the target group.

In our experiments, we extract the group-specific attributes from answers from eleven Stack Exchange communities (e.g., philosophers or game developers), finding high correlations that support the adequacy of the attributes. We assess the effectiveness of our approach against prompting and state-of-the-art steering baselines, focusing on the specificity and factuality of the generated explanations. In addition, we manually evaluate steering success, plausibility and helpfulness in a human study with nine experts from biology, philosophy, and game development. While the results show a tradeoff between specificity and factuality, our approach achieves the best balance between these two criteria, generating significantly better tailored explanations while largely maintaining factuality.

To summarize, our main contributions are:
\begin{enumerate}
\setlength{\itemsep}{0pt}
\item An unsupervised method to obtain style and knowledge attributes of a specific group.
\item A training-free method to enable attribute-based activation steering of LLMs for group-specific explanation generation.
\item Empirical evidence for the effectiveness of respective steering from an expert evaluation.
\end{enumerate}

\section{Related Work}
\label{sec:relatedwork}

\hwfigure{approach-final}{Our approach to creating an attribute-based steering vector $\mathbf{s}_g^{(i)}$ for layer $i$: From explanations of the target group $g$, the $m$ most important style and knowledge attributes are determined. One activation vector is extracted per attribute. These vectors form $\mathbf{s}_g^{(i)}$ which is used to steer an LLM for group-specific explanation generation.}

With the increasing need of explainability in AI, explanations have received attention in computational research \cite{gilpin:2018}. \citet{miller2019whathowwhy} pointed out that explanations are always affected by both the \emph{explainer} as well as the \emph{explainee}.

Therefore, ideally, an explanation should be personalized towards the specific explainee \cite{Sokol2020,Rong2024HumanCenteredXAI}.
The generation of personalized explanations may seem like an instance of text style transfer \cite{jin:2022,reif:2022} or content transfer \cite{ prabhumoye:2019,chen:2024}, where the goal is to induce new content while maintaining style. But, it involves both adjusting the explanations to the explanatory style \textit{and} to the background of the explainee \cite{Rong2024HumanCenteredXAI}.

Since individual preferences are not always available, one alternative is to adapt the explanations to a group of explainees \cite{schneider2019personalized}.
Several works provide human-written explanations \cite{camburu2018eSNLI,fan:2019,rajani2019cosE, wang2020comVE,aggarwal2021ECQA,alshomary:2024}, but none of them include personal information about the explainees.
Thus, recent works investigate how to identify the preferences of specific groups. \newcite{siskou2025fingerpringtprompting} rely on manually designed rhetorical features, limiting scalability though. \newcite{cunningham2023sparseautoencoder} demonstrate that sparse autoencoders can recover interpretable latent features from LLM activations. However, these latent features require semantic labeling to obtain explicit natural-language attributes. Most closely related to our approach, \newcite{patel2023} propose a prompt-based method that automatically extracts style-related attributes from text. Building on this approach, we can also extract knowledge-related~attributes.

The given task of group-specific explanation generation resembles the idea of belief-based argument generation \cite{alshomary:2021} on a certain abstraction level. The authors use plug-and-play language models \cite{dathathri2020PPLM} to adjust arguments to a target audience, which is based only on a specified vocabulary though. Fine-grained personalization is often investigated under the idea of persona prompting \cite{lee:2023}.
For example, in works closely related to our task, LLMs are prompted to tailor explanations to individual user preferences \cite{jeck25PersonalizedExpl,Zhang2025PersonalizedExplGen}.
However, \newcite{rooein2023knowYourAudience} observed that LLMs do not adapt to different age levels, when prompting them accordingly, such as ``Answer this question for 6th graders, what is gravity?''. Moreover, \newcite{siskou2025fingerpringtprompting} found that prompting is not suitable for fine-grained steering as in our setting.
In contrast, \newcite{meng2024attributecontrol} propose attribute-controlled fine-tuning through constrained training objectives and \citet{lai2024styleneurons} identify style-specific neurons to enable target style transfer. However, both methods are primarily evaluated on attributes with clear lexical indicators such as toxicity. Lately, several approaches use continuous user embeddings by encoding style, content, and preferences into dense representations to personalize LLM generation \cite{doddapaneni2024userEmbeddingModel,huber2025userembedding}. While potentially effective in general, these representations make it difficult to independently control specific~attributes.

Recent works employ activation steering \cite{turner2023actisteering, konen2024steeringstyle,chen2025personavector} which assumes that steering vectors can be added to the activations during inference to steer the LLM's output. For multi-attribute steering, \citet{wanfg2025adaptiveSteering} extract steering vectors for different hallucination categories and use trained probes to adapt the steering strength at inference time. \citet{nguyen2025MultiAttribute} train attribute-specific gating functions to enable token-level steering, while \citet{oozeer2025MultiAttribute} train a classifier to derive multi-attribute steering vectors. In contrast, we propose a method that enables multi-attribute steering \emph{without any need for training} as explained in the next section.

\section{Approach}
\label{sec:approach}

In this section, we present our approach to steer an LLM to generate natural language explanations that are tailored to a target group of explainees, say, to game developers or to philosophers. To this end, it models the group's abilities and background in terms of explanatory style and knowledge \cite{jin:2022,reif:2022,alshomary:2021}. 

In the following, we detail the three main steps of our approach as illustrated in Figure~\ref{approach-final}: (1) extracting style and knowledge attributes of a target group from explanatory texts written by the group; (2) creating attribute-specific activation vectors; and (3) computing attribute-based steering vectors to steer an LLM to generate group-specific explanations.

\subsection{Group-Specific Attributes}
\label{sec:approachAttributeVector}

Given a set $G$ of target groups, we identify their most important \textit{style attributes} and \textit{knowledge attributes} from text samples written by members of these groups with a two-stage prompting process. 

For the style attributes, we first reuse the 92 prompts of \citet{patel2023} to let an LLM generate style descriptions of the texts (see \ref{appendix:style-attributes} for details). 
Then, each of these descriptions is rewritten into a list of style attributes (e.g., \textit{The author includes code snippets.}). For the knowledge attributes, we create a new set of 18 prompts to extract six common types of knowledge: factual, conceptual, procedural, metacognitive, situational, and conventional \cite{bloom1956knowledgetaxonomy,deJong1996knowledgetaxonomy,anderson2001knowlegdetaxonomy}.
In addition, we create a new prompt to rewrite each knowledge-related description into a list of knowledge attributes (all prompts are found in \ref{appendix:knowledge-attributes}). Running the two-stage prompting with these prompts, we obtain lists of knowledge attributes (e.g., \textit{The author understands game development.}).

\paragraph{Attribute Selection}
For noise reduction, we perform an LLM-based filtering by classifying each extracted attribute into one of the four classes: \emph{style}, \emph{knowledge}, \emph{both}, or \emph{none} (details in \ref{appendix:llm-based-filtering}). We keep only those attributes that either belong to style or to knowledge. This results in candidate style attributes and candidate knowledge attributes.

On this basis, we select the same set of style attributes $F^{(S)}$ and the same set of knowledge attributes $F^{(K)}$ across all target groups~$G$ to enable comparison across the groups.
To this end, we first cluster semantically-similar candidate attributes, namely using radius-based neighbor clustering using a predefined cosine similarity threshold (see \ref{appendix:clustering} for details).
The rationale for the clustering is that an attribute not appearing frequently might not only indicate that the corresponding style or knowledge is rarely used to describe the groups, but also that the LLM exhibits high syntactic variability when describing it.
Moreover, to avoid too general attributes, we only keep clusters including attributes that were extracted for a maximum number of groups. To ensure that the attributes are not too specific, we additionally require a minimum frequency based on how often they were extracted (see Appendix \ref{appendix:frequency-filtering} for details).

Finally, the attributes are selected in descending order of their frequency, separately for style and knowledge clusters. Any attribute that is too similar to a previously selected one is rejected. Altogether, the outlined process results in the final set of attributes $F = \{f_1, f_2, ..., f_n\}$, where each $f_j \in F^{(S)} \cup F^{(K)}$.
We do not enforce a fixed number $n$, but instead retain all clusters that remain.

\paragraph{Group-Specific Attribute Vectors}
Given a text written by a member of a specific group~$g \in G$, we define its attribute vector $\mathbf{v}_g \in \{0,1\}^{n}$ where the value $v_j = 1$ means that attribute~$f_j$ was predicted for the text during the two-stage prompting process. Given a set of texts, we define the mean $\bar{\mathbf{v}}_g \in [0,1]^{n}$ to be the style and knowledge attribute profile of the group. In addition, we identify the most important style and knowledge attributes~of~$g$.

\subsection{Attribute-Specific Activation Vectors}
\label{sec:approachSteering}
The key idea of our approach is to use the extracted attributes to steer an LLM for group-specific explanation generation, for which we rely on activation engineering \cite{subramani:2022,turner2023actisteering,konen2024steeringstyle}. Activation engineering assumes that the information required to generate a target output is already embedded within the hidden layers of an LLM. Thus, so-called activation-based steering vectors can be added to the activations of specific layers during inference to steer the LLM's output.
In contrast to \citet{konen2024steeringstyle}, we do not extract the activation vectors by processing entire group-specific texts, but we directly use the natural language attributes $f_j \in F$ to enable more fine-grained control. In particular, we extract the activation vectors for the $m$~most important style attributes $F^{(S)}_{m_g} \subset F$ and the $m$ most important knowledge attributes $F^{(K)}_{m_g} \subset F$ identified for a target group $g$.
However, before, each attribute $f_{j} \in F^{(S)}_{m_g} \cup F^{(K)}_{m_g}$ is rewritten as follows: Style attributes are expressed in the second-person perspective (e.g., \textit{You include code snippets.}), as they function as instruction to the LLM. In contrast, knowledge attributes are expressed in the first-person perspective (e.g., \textit{I understand game development.}), reflecting how users typically communicate such information to an LLM.

Via a forward pass, we then extract the activation vector~$\mathbf{a}_{j}^{+(i)}$ based on the rewritten attributes for each layer $i$. In addition, we also extract the activation vectors~$\mathbf{{a}}_{j}^{-(i)}$ for the negated attribute (e.g., \textit{You do NOT include code snippets.}). The final attribute-specific activation vector is based on the difference~\cite{turner2023actisteering}: $\mathbf{a}_{j}^{(i)} \coloneqq \mathbf{a}_{j}^{+(i)}-\mathbf{{a}}_{j}^{-(i)}$.

\subsection{Attribute-Based Steering Vector}
To apply a fine-grained steering based on the $m$~most important attributes $f_{j} \in F^{(S)}_{m_g} \cup F^{(K)}_{m_g}$ of any group~$g \in G$, we introduce the new notion of \emph{attribute-based} steering vectors~$\mathbf{s}_{g}^{(i)}$: We define $\mathbf{s}_{g}^{(i)}$ as the sum of the attribute-specific activation vectors~$\mathbf{a}_{j}^{(i)}$ of layer~$i$.  As not all attributes might be equally important for group $g$, we apply a weighting based on the softmax of the attribute profile~$\bar{\mathbf{v}}_g$:
\begin{eqnarray}
	\mathbf{s}_{g}^{(i)}  \coloneqq & \sum_{f_{j} \in F^{(S)}_{m_g} \cup F^{(K)}_{m_g}}^{} \textrm{softmax}(\bar{\mathbf{v}}_g)_j \cdot \mathbf{a}_{j}^{(i)}
	\label{eq:steeringvector}
\end{eqnarray}

Given an input $x$ (e.g., ``How does DNA work?''), we steer an LLM to generate group-specific explanations by adding $\mathbf{s}_{g}^{(i)}$ to the original activation vector $\mathbf{a}^{(i)}$ for layer $i$ during inference. The~$\lambda$ controls the strength of the steering \cite{konen2024steeringstyle}: 
\begin{eqnarray}
	\hat{\mathbf{a}}^{(i)}(x) & \coloneqq & \mathbf{a}^{(i)}(x) +  \lambda \cdot \mathbf{s}_{g}^{(i)}
\end{eqnarray}

\section{Data}

This section details the data utilized in our experiments to extract group-specific attributes and to generate group-specific explanations.

\subsection{Group-Specific Attribute Extraction} 
\label{sec:datasetsAttributeExtracation}
To extract group-specific attributes, we only need data containing texts written by members of specific groups. We rely on Stack Exchange which is a collection of Q\&A forums, each covering a specific domain (e.g., game development). For our purposes, we make the simplifying assumption that the answers in a forum of a specific domain are provided by members of the respective group (e.g.,~game developers). While this assumption will not always hold true, it seems reasonable in general according to our inspection of several samples. We randomly selected 6,000 answers from each of the eleven groups seen in Figure~\ref{fig:profile-correlations} (details in \ref{appendix:stackexchange}).

\subsection{Group-Specific Explanation Generation}
\label{sec:datasetsExplanationGeneration}

For group-specific explanations generation, explanatory questions are needed that are general enough to plausibly be asked by diverse groups. To this end, two types of questions are employed:

\paragraph{ELI5 Questions} 
The \textit{Explain Like I'm 5} (ELI5) subreddit \cite{fan:2019} is a Q\&A forum on Reddit providing simplified explanations. We use the questions of the ELI5 validation split (1,507) and test split (600) from the KILT benchmark \cite{petroni2021kilt} for our steering experiments. Since the answers are written for a general audience, they cannot serve as a ground-truth for our experiments.

\paragraph{Science Questions} 
\citet{rooein2023knowYourAudience} created a set of scientific questions to investigate whether LLMs can adapt their answers to different age and education levels. We adopt these questions and filter for duplicates resulting in 97 further questions.

\section{Experiments}

We carried out two main experiments to evaluate the adequacy of the extracted attributes and the steering effectiveness of our approach.\footnote{The code and data can be found under \url{https://github.com/webis-de/EMNLP-26}.}

\subsection{Group-Specific Attribute Extraction}
\label{sec:experimentAttributeVector}

We run the two-stage prompting process of our approach to identify the most important attributes of each target group $g \in G$ with $|G|=11$.

\paragraph{Models} 

For prompting, we employed the LLM \texttt{Qwen2.5-7B-Instruct} \cite{yang2024qwen2.5}. The attribute filtering was performed using the bigger version \texttt{Qwen2.5-72B-Instruct}. To obtain vector representations of each attribute, we used \texttt{Qwen3-Embedding-8B} \cite{zhang2025qwen3embedding}.

\paragraph{Experimental Setup}
From the Stack Exchange data, we randomly sample 500 answers per group $g$ and run the two-stage prompting process, the LLM-based filtering and the clustering to then remain with $|F| = 1250$ attributes including 340 style and 910 knowledge attributes (see \ref{appendix:attribute-extracation} for details). 

Finally, we create the group-specific attribute vectors $\mathbf{v}_g \in \{0,1\}^{1250}$ for each of the 500 answers per group and compute the group profiles~$\bar{\mathbf{v}}_g$. By computing the point-biserial correlation \cite{lev1949PointBiserialCorrelation} for all $\mathbf{v}_g$, we identify the $m$~most important style and the $m$ most important knowledge attributes per group (see \ref{appendix:important-attributes} for~details). To assess the impact of $m$ on the steering effectiveness, we consider $m \in \{3, 5, 10\}$. These rather small numbers are motivated by the observation that even humans struggle with considering too many attributes simultaneously \cite{miller1956cognitiveCapacity}.

To investigate the adequacy of the attributes, we compute the Pearson correlation coefficient~\cite{pearson1896} of the profiles $\bar{\mathbf{v}}_g$ based on the attribute vectors predicted by the model described in Appendix~\ref{appendix:group-score} for the 6000 answers per group $g$.

\subsection{Group-Specific Explanation Generation}

Given the most important attributes of a specific group $g \in G$, we steer an LLM to generate group-specific explanations. We evaluate our approach against two baselines: \emph{prompting} and standard \emph{activation-based steering} \cite{konen2024steeringstyle}.

\paragraph{Models} 

For our steering experiments, we evaluate three model families of comparable size to ensure a fair comparison: \texttt{Qwen2.5-7B-Instruct}, \texttt{Llama-3.1-8B-Instruct} \cite{grattafiori:2024} and \texttt{Ministral-8B-Instruct-2410}\footnote{Model Card, \url{https://huggingface.co/mistralai/Ministral-8B-Instruct-2410}}.
To further examine how model capacity influences steering effectiveness, we additionally include a larger variant from the Qwen family (\texttt{Qwen2.5-32B-Instruct}).

\paragraph{Experimental Setup}

For the automatic hyperparameter tuning of all activation-based steering methods, we randomly sampled 100 questions from the val. split of the ELI5 dataset (same questions for all methods). Based on the best hyperparameters, the final experiment was conducted on the 600 ELI5 test questions and the 97 science~questions.

For the prompting baseline, we designed two system prompts (full prompts in Appendix~\ref{appendix:prompting-baseline}). The first prompt includes only the target group (called \emph{group} below). The second prompt additionally includes the $m$~most important attributes for that group (\emph{group~+~attributes ($m$)}). In all cases, we  instructed the LLM to limit its explanations to 200 words as LLMs tend to generate overly long explanations which differs from how humans usually explain \cite{miller2019whathowwhy}. For example, the Stack Exchange data reveals an average length of 168~words.

The activation-based steering approaches utilized the same prompts. In addition, the steering vector $\mathbf{s}_{g}^{(i)}$ is added to the activations at each layer~$i$ of a selected set of layers~$I$. As \citet{konen2024steeringstyle}, we select the best performing $I$ from all sets of layers $I_0=\{0, 1, 2\}, I_1=\{1, 2, 3\}, \ldots$ under a~sliding window of size $3$ on the validation set. We select the best $\lambda$ from  $\{0.25, 0.5, \ldots, 2.5\}$ and report the selected factor $\lambda$ and layers~$I$ below.

For the activation-based steering baseline \cite{konen2024steeringstyle}, the steering vector is computed by averaging the activation vectors extracted for the 500 Stack Exchange answers of the target group~$g$ and subtracting the mean activation vector of all other groups $G \setminus g$.
For our approach, we construct the steering vectors according to Equation~\ref{eq:steeringvector} based on the $m$~most important attributes of group $g$.

\paragraph{Evaluation Metrics}

We automatically evaluate the steering effectiveness in terms of the group~\emph{specificity} and the \emph{factuality} of the generated explanations and the harmonic mean as follows:

To assess the group specificity, we train a classifier $c: [0,1]^{1250} \rightarrow \mathbb{R}^{11}$ that takes an attribute vector~$\mathbf{v} \in [0,1]^{1250}$ as input and predicts a score for every possible target group. We use this metric also for the hyperparameter tuning (details in \ref{appendix:group-score}).  Given $\mathbf{v}$ for an explanation tailored to group $g$, the specificity score in $[0, 1]$ is obtained by applying the softmax to $c(\mathbf{v})$ and selecting the score corresponding to $g$. The score can be interpreted as the probability how much the explanation is tailored: 
\begin{eqnarray}
	\textit{specificity} & \coloneqq & \textrm{softmax}(c(\mathbf{v}))_g 
\end{eqnarray}

For factuality, we employ the \textsc{FActScore} metric \cite{min2023factscore}, which computes the probability of atomic facts $X$ extracted from input text that are supported by a knowledge source (see \ref{appendix:factscore} for details).
\begin{eqnarray}
	\textit{factuality} & \coloneqq & \textsc{FActScore}(X)
\end{eqnarray}

\noindent
The metric requires specifying a topic that maps to a Wikipedia title. We thus report results only for the science questions, as part of the ELI5 questions cannot be mapped to such a title, for example, ``Why was quicksand such a common film trope when it's not a problem [...] in real life?''

Since high-quality explanations should be both well understandable by the explainee and factually accurate \cite{loefstroem2022evalcriteria}, we also consider the harmonic mean to capture the balance between the evaluation metrics (see \ref{appendix:hmean} for details).

\section{Results and Discussion}

In this section, we first analyze the adequacy of the extracted style and knowledge attributes. Then, we present the findings on the steering effectiveness.

\subsection{Attribute Profile Correlation}

\bsfigure[width=0.48\textwidth]{group-attribute-correlation-num.ai}{Pearson correlation of attribute profiles in terms of style (lower triangle, white text) and knowledge (upper triangle, black text) for the 11 target groups. Bold values denote statistical significance~($p < .05$). \label{fig:profile-correlations}}

\begin{table*}[t]
	\small
	\centering
	\renewcommand{\arraystretch}{0.95}
	\setlength{\tabcolsep}{5pt}
	\begin{tabular}{lllrlrrr}
		\toprule
		\textbf{LLM} & \textbf{Prompt Components} & \textbf{Steering Vectors} &
		\textbf{Factor} $\mathbf{\lambda}$ & \textbf{Layers $I$} &
		\textbf{Specificity $\uparrow$} & \textbf{Factuality $\uparrow$} & \textbf{HMean $\uparrow$} \\
		\midrule
		\multirow{6}{*}{\rotatebox{90}{Qwen 7B}}
		& Group & - & - & - & 0.128 & \textbf{0.322} & 0.183\\
		& Group \textbf{+ attributes (5)} & - & - & - & 0.224 & \underline{0.303} & 0.257\\
		& Group & \citet{konen2024steeringstyle} & 1.75 & [17,18,19] & \underline{0.694} & 0.126 & 0.213\\
		& Group \textbf{+ attributes (5)} & \citet{konen2024steeringstyle} & 1.50 & [18,19,20] & \textbf{0.803} & 0.096 & 0.171\\
		& Group & \textbf{Our approach (5)} & 2.00 & [10,11,12] & \textsuperscript{$\dagger$}0.386 & \textsuperscript{$\ddagger$}0.264 & \textbf{0.313}\\
		& Group \textbf{+ attributes (5)} & \textbf{Our approach (5)} & 1.50 & [10,11,12] & \textsuperscript{$\dagger$}0.442 & \textsuperscript{$\ddagger$}0.225 & \underline{0.299}\\
		\midrule
		\multirow{6}{*}{\rotatebox{90}{Llama 8B}}
		& Group & - & - & - & 0.217 & \textbf{0.326} & 0.261\\
		& Group \textbf{+ attributes (5)} & - & - & - & 0.410 & 0.258 & \textbf{0.316}\\
		& Group & \citet{konen2024steeringstyle} & 1.25 & [15,16,17] & \underline{0.532} & 0.195 & 0.285\\
		& Group \textbf{+ attributes (5)} & \citet{konen2024steeringstyle} & 1.00 & [17,18,19] & \textbf{0.664} & 0.149 & 0.244\\
		& Group & \textbf{Our approach (5)} & 0.75 & [17,18,19] & \textsuperscript{$\dagger$}0.300 & \textsuperscript{$\ddagger$}\underline{0.279} & 0.289\\
		& Group \textbf{+ attributes (5)} & \textbf{Our approach (5)} & 0.75 & [17,18,19] & \textsuperscript{$\dagger$}0.448 & \textsuperscript{$\ddagger$}0.215 & \underline{0.291}\\
		\midrule
		\multirow{6}{*}{\rotatebox{90}{Ministral 8B}}
		& Group & - & - & - & 0.111 & \textbf{0.376} & 0.171 \\
		& Group \textbf{+ attributes (5)} & - & - & - & 0.191 & 0.346 & 0.246 \\
		& Group & \citet{konen2024steeringstyle} & 1.00 & [20,21,22] & \underline{0.300} & 0.252 & \underline{0.274} \\
		& Group \textbf{+ attributes (5)} & \citet{konen2024steeringstyle} & 1.00 & [21,22,23] & \textbf{0.511} & 0.187 &  \textbf{0.274}\\
		& Group & \textbf{Our approach (5)} & 1.00 & [5,6,7] & \textsuperscript{$\dagger$}0.128 & \textsuperscript{$\ddagger$}\underline{0.354} & 0.188 \\
		& Group \textbf{+ attributes (5)} & \textbf{Our approach (5)} & 1.00 & [5,6,7] & \textsuperscript{$\dagger$}0.215 & \textsuperscript{$\ddagger$}0.330 & 0.260 \\
		\bottomrule
	\end{tabular}
	\caption{
	Main automatic evaluation results: \emph{Specificity}, \emph{factuality}, and \emph{harmonic mean} of all tested \emph{LLMs}, \emph{prompt components}, and \emph{steering vectors} on the science questions \citep{rooein2023knowYourAudience}, along with steering \emph{factor} $\lambda$ and steered \emph{layers} $I$. Bold components/vectors are part of our approach. Best values per LLM bold, second underlined. \textsuperscript{$\dagger$} and \textsuperscript{$\ddagger$} denote significant improvements over prompting and \citet{konen2024steeringstyle} per prompt respectively ($p < .05$)
	}
	\label{tab:steering_eval_science_questions}
\end{table*}

To investigate the adequacy of the extracted attributes, Figure~\ref{fig:profile-correlations} presents a joint view of how the explanatory style profiles (lower triangle) and the knowledge profiles (upper triangle) of the eleven target groups correlate in terms of Pearson's $r$.

Style correlations are generally high across scientific and technical groups such as \emph{biologists} and \emph{electrical engineers}. The knowledge profiles are more differentiated: Some scientific groups, such as \emph{chemists} and \emph{physicists}, also show a high correlation, while others display weaker correlations despite similar style (e.g., \emph{biologists} and \emph{physicists}). \emph{Computer scientists}, \emph{software engineers}, and \emph{game developers} have a weaker knowledge overlap with natural scientists. The non-technical groups (\emph{historians}, \emph{philosophers}, \emph{politicians}) form a distinct cluster across both style and knowledge.

Overall, these results suggest that explanatory style is more homogeneous across all groups, indicating a shared way of structuring explanations. In contrast, knowledge profiles are more diverse, and non-technical groups remain distinct from all other groups. Thus, the created group profiles $\bar{\mathbf{v}}_g$ capture both expected shared explanatory style and meaningful differences in domain-specific knowledge.

\begin{table*}[t]
	\small
	\centering
	\renewcommand{\arraystretch}{0.95}
	\setlength{\tabcolsep}{5pt}
	\begin{tabular}{lllrrrr}
		\toprule
		&\textbf{Prompt Components} & \textbf{Steering Vectors} & \textbf{Steering Success $\uparrow$} & \textbf{Plausibility $\uparrow$} & \textbf{Helpfulness $\uparrow$} & \textbf{HMean $\uparrow$}\\
		\midrule
		(a) &Group & - & 3.12$\;\pm$1.21 & \textbf{4.56}$\;\pm$0.86 & \textbf{4.03}$\;\pm$1.10 & 3.81  \\
		(b) & Group & \citet{konen2024steeringstyle} & 3.46$\;\pm$1.44 & 2.75$\;\pm$1.46 & 2.70$\;\pm$1.36 & 2.93\\
		(c) & \textbf{Group + attributes (5)} & \textbf{Our approach (5)} & \textsuperscript{$\dagger \ddagger$}\textbf{4.17}$\;\pm$1.01 & \textsuperscript{$\ddagger$}3.87$\;\pm$1.18 & \textsuperscript{$\ddagger$}3.64$\;\pm$1.17 & \textbf{3.88} \\
		\bottomrule
	\end{tabular}
	\caption{
	Manual evaluation results: \textit{Steering success}, \textit{plausibility}, and \textit{helpfulness} of the Qwen 7B explanations generated by our approach and two baselines, averaged over 50 randomly-sampled questions, 25 per dataset. Best values bold. \textsuperscript{$\dagger$} and \textsuperscript{$\ddagger$} denote significant improvements over prompting and \citet{konen2024steeringstyle} respectively ($p < .05$).
	}
	\label{tab:human_study_both}
\end{table*}

\subsection{Automatic Evaluation of Explanations}

Table~\ref{tab:steering_eval_science_questions} presents the average results on the 97 science questions for three model families, averaged over all eleven target groups, along with the harmonic mean (HMean). Since the specificity results for the ELI5 questions exhibit similar patterns, we outsource them to Appendix~\ref{appendix:results-eli5}. Significant improvements over the baselines for the two criteria are determined using the Wilcoxon signed-rank test for each prompt component respectively. 

Table~\ref{tab:steering_eval_science_questions} shows including our extracted attributes in the prompt improves the group specificity across all approaches and models. However, this increase in specificity results in lower factuality. 
Our approach generally seems the best midway between specificity and factuality, in case of Qwen also achieving the highest harmonic means (0.313 and 0.299) and the second best value with 0.291 for Llama. 
Depending on the model, the highest specificity ranges from 0.511 to 0.803, all for \citet{konen2024steeringstyle} with our extracted attributes in the prompt. For Ministral, this even leads to best harmonic mean (0.274). The best factuality for all models is achieved by the prompting baseline (0.322--0.376), which is expected given its lower~specificity.

The ablation results in Appendix~\ref{appendix:results-ablations} show that for $m \in \{3, 10\}$ there are~no performance differences, while a bigger model size can achieve a better balance between specificity~and factuality.

In line with prior works \cite{konen2024steeringstyle,bogdan2025middlelayers}, the best-performing layers for all models tend to be around the middle of their respective layer counts. For Ministral, however, the rather early layers~$I=[5,6,7]$ out of 36 layers were selected for our approach which may explain its limited performance. While we expect more exhaustive hyperparameter tuning to help, we leave this for future~work.

Overall, the results indicate a tradeoff between generating explanations that are well tailored to a target group and maintaining factual precision. As an example, a use of analogies as in Figure~\ref{fig:sample} can make explanations more tailored to the target group. However, this increased specificity can lead to decreased factual accuracy \cite{wiley2018analogyfactuality}: While the method of \citet{konen2024steeringstyle} tends to ``oversteer'', resulting in high specificity but low factuality, our approach provides the best balance between the two criteria, which particularly turns out effective for Qwen. This raises the question of whether the effectiveness of steering for this model is affected by using Qwen for both attribute extraction and explanation generation. To investigate this, we conduct an ablation study in which Llama and Ministral are used for attribute extraction instead. The results show that our approach remains robust across different attribute extraction models, suggesting that its effectiveness does not depend on using Qwen (see Appendix \ref{appendix:ablation-study-attribute-extraction} for details).

\bsfigure[width=0.48\textwidth]{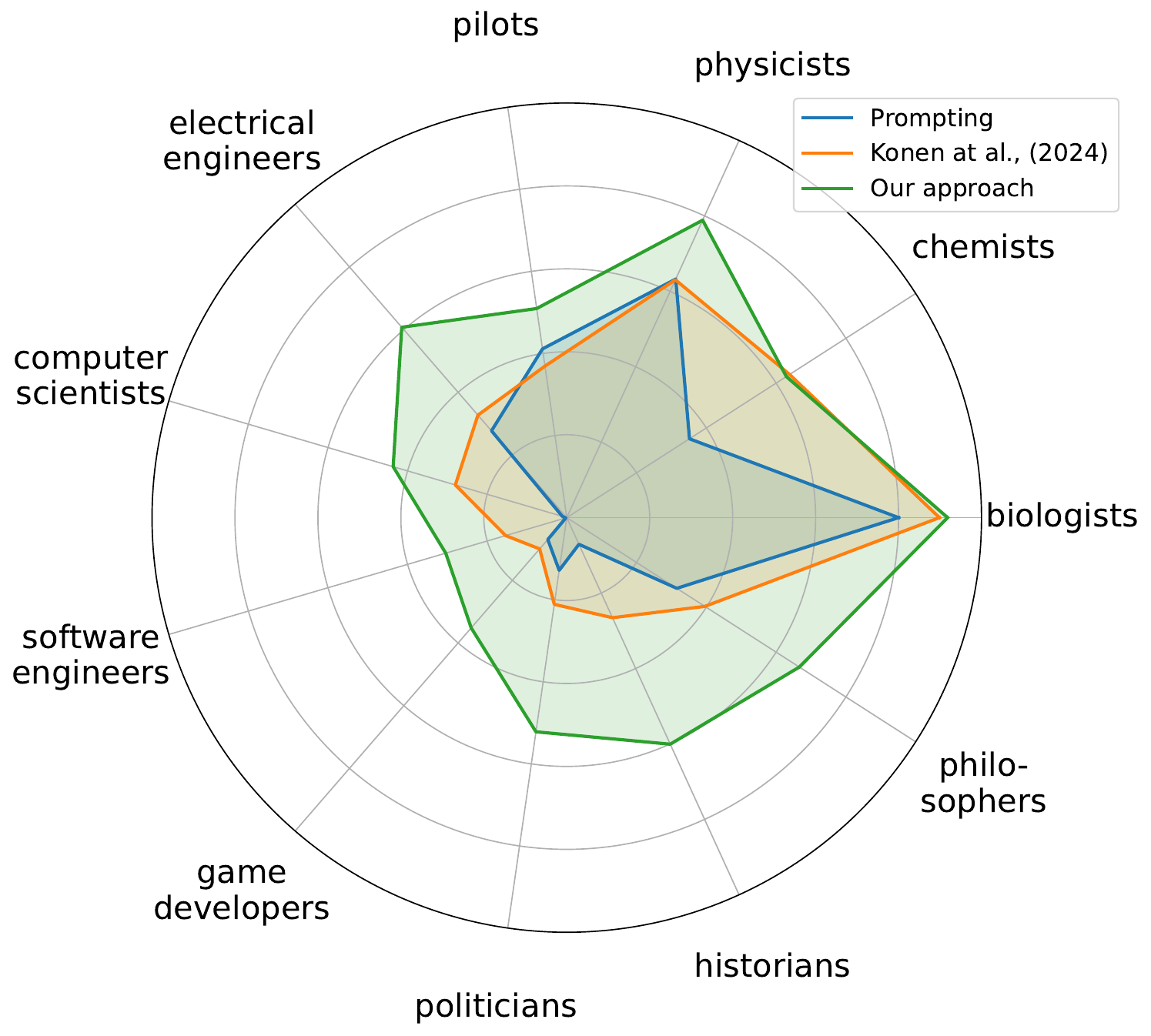}{Per-group harmonic mean of \emph{specificity} and \emph{factuality} of the explanations generated by Qwen (7B) for the 97 science questions using the prompting baseline, the method of \citet{konen2024steeringstyle} and our approach. The more outside a point, the higher the harmonic mean.
\label{fig:hmean-per-group-science-questions}}

Figure~\ref{fig:hmean-per-group-science-questions} shows the per-group harmonic mean based on the specificity and factuality of the explanations generated by Qwen on the 97 science questions. Blue represents the prompting baseline, orange the method of \citet{konen2024steeringstyle}, and green our approach.
All methods include the target group in the prompt, while our approach additionally uses the five most important attributes. The prompting baseline performs well for physicists and biologists but the harmonic mean drops to near zero for computer scientists. Our approach achieves the best harmonic mean for nearly all groups, particularly for scientific and non-technical groups.

\begin{table*}[t]
	\centering
	\small
	\begin{tabular}{@{}l|p{15.2cm}@{}}
		\toprule
		 & \textbf{Question: How does DNA work?} \textbf{Target group:  Game developers}\\
		\midrule
		
		\multirow{7}{*}{\rotatebox{90}{Prompting}}
		& DNA, or deoxyribonucleic acid, is the molecule that carries genetic information in living organisms. It's structured as a double helix, with two strands coiled around each other. Each strand is made of nucleotides, which contain a sugar, a phosphate group, and one of four nitrogenous bases: adenine (A), thymine (T), cytosine (C), and guanine (G). The sequence of these bases along the DNA strand encodes instructions for making proteins, which are essential for life. The base pairs (A-T and C-G) hold the two strands together. When cells divide, enzymes read the DNA sequence, transcribe it into RNA, and then translate the RNA into proteins. \textbf{In games, you might use DNA analogies for character customization, evolution systems, or genetic traits that affect gameplay mechanics}.\\
		\midrule
		
\multirow{7}{*}{\rotatebox{90}{\citet{konen2024steeringstyle}\phantom{i}}}
		& %
			Sure! Here’s a simple way to set up your DNA system:  [\ldots]
			\begin{verbatim}
				public class DNA {
					private List<Vector2> points;
					public DNA() {
						points = new List<Vector2();
					}
				[...]
			\end{verbatim} 
			\vspace{-1.4em}
		\\
		\midrule
		
		\multirow{6}{*}{\rotatebox{90}{Our approach}}
		&  From a game developer's perspective, you might want to think of \textbf{DNA as a blueprint that defines the structure and behavior of your game}. Just like DNA codes for the traits of an organism, your \textbf{game's code defines its behavior and features}. For instance, you might have functions that initialize game objects, handle user input, or manage game states. These \textbf{functions are like the genes that determine how your game functions at different stages}. You could also \textbf{consider using comments to explain complex parts of your code, much like DNA sequences are annotated to provide context}. This helps in making the code more understandable and maintainable, enhancing the overall gaming experience.
		\\
		\midrule
		
	\end{tabular}
	\caption{Explanation samples of the two baselines (prompting and \citet{konen2024steeringstyle}) and our approach generated by Qwen (7B) for the question ``How does DNA work?'' tailored to \emph{game developers}. The two baselines use the target group in the prompt and our approach additionally uses the five most important attributes in the prompt.\label{tab:samples-1-approach} \label{tab:samples-1-all-approaches}}
\end{table*}

\subsection{Human Evaluation of Explanations}

To manually evaluate the steering effectiveness of our approach, we conducted a user study with nine experts (three biologists, three game developers, and three philosophers), hired on Upwork.
We randomly sampled 25 science questions \cite{rooein2023knowYourAudience} and 25 ELI5 questions. For each question, the experts are presented only with explanations tailored to their own target group. We compared three approaches based on Qwen~7B: (a)~the prompting baseline with the \emph{group} prompt (representing an out-of-the-box LLM); (b)~the approach of \citet{konen2024steeringstyle} with the \emph{group} prompt (representing state-of-the-art steering); and (c)~our approach with the five most important attributes in prompt and activitation, since it seemed most robust in automatic evaluation across all LLMs. We assess the \textit{steering success} and \textit{plausibility} of the generated explanations on a 5-point Likert scale from \emph{not at all} (1) to \emph{fully}~(5). Beyond steering effectiveness, we also assess the \emph{helpfulness} of the explanations on the same scale. More details are found in the~Appendix~\ref{appendix:human-study}.

Table \ref{tab:human_study_both} shows the mean and standard deviations for the three criteria each along with the harmonic mean (HMean). Significant improvements over the baselines are tested with the Wilcoxon signed-rank test. Our approach significantly outperforms the two baselines in terms of steering success (4.17) while the prompting baseline yields the best plausibility values (4.56). This supports the specificity--factuality tradeoff seen in the automatic evaluation. The results for \citet{konen2024steeringstyle} again indicate ``oversteering'' denoted by the lowest helpfulness (2.70) and plausibility (2.75) values. Overall, our approach achieves the best harmonic mean of 3.88.

Manual analysis of the explanations (complete samples in Appendix \ref{appendix:samples}) demonstrates the high steering success of our approach. For example, as shown in Table \ref{tab:samples-1-all-approaches}, it adapts the explanation of DNA using analogies that resonate with game developers. In contrast, the explanation of the prompting baseline is barely tailored to game developers. While the explanation generated by \citet{konen2024steeringstyle} is entirely code-based and thus familiar to game developers, it fails to effectively explain DNA.

Interestingly, higher steering success does not appear to translate into greater helpfulness.
One possible reason may be the inherent subjectivity of the task, as indicated by the relatively low inter-annotator agreement (Krippendorff's $\alpha$ ranging from 0.178 to 0.489).
Moreover, we observe no significant difference in the helpfulness scores for the science questions (see Table~\ref{tab:human_study_rooein} in appendix). We speculate that the complexity of these questions enables more benefit of group-specific explanations, while the rather general topics of the ELI5 questions could already be well understood with generic explanations. We further found that low ratings are partly because of domain shifts without answering the question anymore. Thus, our findings indicate the need for steering methods to also explicitly model the type of the question to ensure both successfully tailored and helpful~explanations.

\smallskip
Overall, both the automatic and manual evaluations indicate that, when generating group-specific explanations using state-of-the-art LLMs, there is a tradeoff between tailoring the explanations to a target group and maintaining factuality/plausibility. Our approach based on the five most important style and knowledge attributes of the target group achieves the best balance between these criteria.

\section{Conclusion}
\label{sec:conclusion}

In this paper, we have presented a three-step approach to realize group-specific explanation generation: (1) automatically identifying the most important attributes in terms of style and knowledge of a specific target group; (2) extracting attribute-specific activation vectors; in order to (3) compute \emph{attribute-based} steering-vectors to steer an LLM for group-specific explanation generation.

We have evaluated the steering effectiveness of our approach for 11 target groups based on automatic metrics and a human study with nine experts from three domains. Our results indicate a tradeoff between generating well-tailored explanations and maintaining factuality. Overall, our approach achieves the best balance between these criteria, providing evidence for our hypothesis that one can learn from explanations \emph{of}~a specific target group how to best explain a topic \emph{for} that group. Thereby, our work contributes to the understanding of generating group-specific explanations computationally. However, we also observed that steering success does not necessarily correlate with the helpfulness of explanations. In future work, we thus seek to better understand what tailoring is most helpful.

\section{Limitations}
\label{sec:limitations}

In this paper, we generate group-specific explanations computationally. While group-specific explanations are a possible alternative when individual characteristics are not available, they can only account in a limited way for those individuals whose abilities or background notably deviate from the mean group profile. However, since our approach relies on predefined attributes, it can easily be extended to steer an LLM based on attributes selected differently, for example, attributes that a user specifies using a slider. 

Furthermore, our experiments are limited to the English language and a fixed set of target groups. Although the results are quite promising, it remains to be studied how well our approach generalizes to other languages and target groups.

In general, our notion of \emph{group} is operationalized via Stack Exchange communities, assuming that contributors to a given forum (e.g., game development) are representative of a target group (e.g., game developers). While this assumption seems reasonable in general, it may not hold universally: Users may participate across domains, vary in expertise, or write answers that do not reflect the typical explanatory style of the respective group.

Beyond that, our attribute extraction process relies on LLM-based prompting. Although this enables scalability and avoids manual feature engineering, it also inherits biases of the underlying model. However, as our evaluation shows, the created group profiles based on the extracted attributes of the 11 groups appear adequate, capturing both: expected shared explanatory style and meaningful differences in domain-specific. Moreover, we maintain that Stack Exchange is a high-quality data source because of its reputation system, which allows the sites to self-moderate. Additionally, since we only consider up-voted answers, we can reasonably assume that contributing users demonstrate expertise of their respective target groups (otherwise, they could not provide such helpful answers). Consequently, group overlap should only minimally impact the adequacy of the extracted attributes.
 
Finally, we assess factuality of explanations based on the \textsc{FActScore} metric \cite{min2023factscore} which also relies on LLMs, namely, for fact validation. Since LLMs often lack factual accuracy due to the hallucination problem \cite{Ji2023Hallucination, wang2023claimcheckviaLLM}, we do not judge absolute factuality values but relative differences. Our evaluation indicates a tradeoff between group specificity and factuality when using state-of-the-art LLMs for group-specific explanation generation. Although our approach achieves the best balance between these two criteria, increased specificity can still introduce oversimplifications that reduce factual precision. We recommend that future methods should focus especially on this tradeoff when generating group-specific explanations.

\section{Ethical Considerations}
\label{sec:ethics}

Since our focus is on group-specific explanations, a primary concern is the risk of stereotyping. Even when derived empirically, the created group profiles may present an overly-simplified view of the target group's knowledge and abilities. Consequently, the profiles may mix heterogeneous subgroups or overlook important variations within a group. In addition, since we extract the group-specific attributes from explanations on Stack Exchange, the generated explanations may reflect existing biases present in the source data.

Moreover, although our paper focuses on generating tailored explanations and explicitly evaluates factuality, there is a risk to misuse our approach in contexts beyond harmless educational settings. For example, it could be used in targeted persuasion or to spread misinformation to specific groups. We cannot fully prevent such misuse, even though it is neither the intended use of our approach, nor is the approach anyhow optimized toward this.

Finally, we acknowledge the limited sample size and potential subjectivity in rating the steering success, helpfulness, and plausibility in our human study. Future studies involving broader and more diverse participants could help to further examine the impact of group-specific explanations.

\section*{Acknowledgments}

This work has been supported by the research project “HybrInt - Hybrid Intelligence through Interpretable AI in Machine Perception and Interaction” (Zukunft Nds, Niedersächsisches Ministerium für Wissenschaft, Grant ID: ZN4219) and by the Deutsche Forschungsgemeinschaft (DFG, German Research Foundation) under project number TRR 318/3 2026 -- 438445824. We thank the anonymous reviewers for their insightful feedback. The writing and implementation were supported by DeepL and ChatGPT but the authors reviewed and, if needed, revised all AI-assisted~content.

\bibliography{emnlp26-attribute-analysis-and-steering}
\appendix
\newpage
\section{Group-Specific Attribute Extraction}
\label{appendix:attribute-extraction}
\subsection{Stack Exchange Dataset}
\label{appendix:stackexchange}
To extract group-specific attributes, we rely on English Stack Exchange data.\footnote{Stack Exchange, \url{https://stackexchange.com}} 
Stack Exchange is a collection of Q\&A forums, each covering a specific domain (e.g., game development or philosophy). We used the official Stack Exchange Data Dump\footnote{Stack Exchange Data Dump, \url{https://archive.org/details/stackexchange_20250630}} of all Stack Exchange domains (as of June 30, 2024) and filter for explanatory questions based on the questions words \textit{what}, \textit{how}, \textit{why} \cite{miller2019whathowwhy}. For our purposes, we make the simplifying assumption that the answers in a forum of a specific domain are provided by members of the respective group (e.g., game developers). Note, our method does not depend on Stack Exchange data specifically. The underlying data can easily be exchanged to enable steering for other groups. We only need texts written by members of a specific group.

To ensure high-quality data, we filtered the answers based on the following criteria: (1)~a minimum length of 50 words, to consider detailed explanations. (2)~a maximum tokenized length of 512 tokens since this is the max length that the model \texttt{deberta-v3-large} can handle (see Appendix \ref{appendix:group-score}), and (3)~at least one upvote, as a minimum guarantee that anyone finds the answer helpful.

Afterwards, to ensure a balanced distribution, we consider only Stack Exchange forums that have at least 6000 answers and randomly sample 6000 answers each from the resulting eleven groups:
\begin{quote}
	\em biologists, chemists, computer scientists, electrical engineers, game developers, historians, philosophers, physicists, \\pilots, politicians, software engineers 
\end{quote}

\subsection{Style Attribute Prompts}
\label{appendix:style-attributes}

\citet{patel2023} implemented an unsupervised method to extract style attributes in order to create style representations of author style in text. For this, they introduce a two-stage prompting. For the first stage, they created six open-ended prompts that generate descriptions of a text on a broad dimension of style. For example, this is the open-ended prompt to get a description of the unique grammar style of the author:

\begin{quotation}
	\noindent
	\small
	\texttt{Write a long paragraph describing the \textbf{unique grammar style} of the following passage without referring to specifics about the topic. \newline \newline Passage: ... \newline \newline Description:}
\end{quotation}

In addition, they created 87 targeted prompts to generate descriptions about specific dimensions of style derived from linguistic and psychological categories \cite{Tausczik2010MeaningofWords}. For example, this is the targeted prompt to get a description if the author uses any figurative language. We added ``If not, keep the answer short.'' since in our experiments the LLM hallucinated otherwise:
\begin{quotation}
	\noindent
	\small
	\texttt{Write a description of whether the author of the following passage has any
		\textbf{figurative language}. If not, keep the answer short. \newline \newline Passage: ... \newline \newline Description:}
\end{quotation}

This results in a total of 93 prompts. We adopt 92 of these prompts (6 open-ended prompts, 86 targeted prompts) since one targeted prompt regarding swear words was created twice. 

For the second stage, \citet{patel2023} created a prompt to rewrite the descriptions to natural language style attributes beginning with ``The author ...''. We added ``These sentences have to be short. Avoid examples. Avoid negations. Use only the information that is present in the description.'' since in our experiments the LLM hallucinated otherwise:
\begin{quotation}
	\noindent
	\small
	\texttt{Here's a description of an author's writing style for a passage: ...\newline \newline Rewrite this description as a long list of short sentences describing the author’s writing style where each sentence is in the format of 'The author is X.' or 'The author uses X.'. These sentences have to be short. Avoid examples. Avoid negations. Use only the information that is present in the description. \newline \newline Output:}
\end{quotation}

After running this two-stage prompting on texts written by 1,000 different authors, they extracted nearly 1.3M style attributes. For the final style vector, they selected only 768 attributes. The first 87 attributes correspond to the style attributes extracted by the 87 targeted prompts. The remaining 681 are downselected from the remaining style attributes based on filtering heuristics (frequency, similarity, interpretability).

\subsection{Knowledge Attribute Prompts}
\label{appendix:knowledge-attributes}

We extend the two-stage prompting process of \citet{patel2023} by creating 18 additional open-ended prompts to extract knowledge attributes from a given explanation. We create prompts to get descriptions about six common types of knowledge \cite{bloom1956knowledgetaxonomy,deJong1996knowledgetaxonomy,anderson2001knowlegdetaxonomy}.

For each type of knowledge, we create a general prompt. For example, this is the general prompt to extract the factual knowledge of the author:

\begin{quotation}
	\noindent
	\small
	\texttt{Write a long paragraph describing the \textbf{factual knowledge} presented in the following passage.}
\end{quotation}
In addition, we create more specific prompts related to this type of knowledge:

\begin{quotation}
	\noindent
	\small
	\texttt{Write a long paragraph describing the \textbf{terminology} presented in the following passage.}
\end{quotation}

\begin{quotation}
	\noindent
	\small
	\texttt{Write a long paragraph describing the \textbf{specific facts} presented in the following passage.}
\end{quotation}

We use the following prompts to extract the conceptual knowledge of the author:
\begin{quotation}
	\noindent
	\small
	\texttt{Write a long paragraph describing the \textbf{conceptual knowledge presented} in the following passage.}
\end{quotation}

\begin{quotation}
	\noindent
	\small
	\texttt{Write a long paragraph describing the \textbf{classifications and categories} presented in the following passage.}
\end{quotation}

\begin{quotation}
	\noindent
	\small
	\texttt{Write a long paragraph describing the \textbf{principles and generalizations} presented in the following passage.}
\end{quotation}

\begin{quotation}
	\noindent
	\small
	\texttt{Write a long paragraph describing the \textbf{theories, models, and structures} presented in the following passage.}
\end{quotation}

\begin{quotation}
	\noindent
	\small
	\texttt{Write a long paragraph describing the \textbf{trends and sequences} presented in the following passage.}
\end{quotation}

To extract the procedural knowledge, we utilize the following prompts:
\begin{quotation}
	\noindent
	\small
	\texttt{Write a long paragraph describing the \textbf{procedural knowledge} presented in the following passage.}
\end{quotation}

\begin{quotation}
	\noindent
	\small
	\texttt{Write a long paragraph describing the \textbf{subject-specific skills and algorithms} presented in the following passage.}
\end{quotation}

\begin{quotation}
	\noindent
	\small
	\texttt{Write a long paragraph describing the \textbf{subject-specific techniques and methods} presented in the following passage.}
\end{quotation}

\begin{quotation}
	\noindent
	\small
	\texttt{Write a long paragraph describing the \textbf{criteria for selecting and applying appropriate procedures} presented in the following passage.}
\end{quotation}

To extract the metacognitive knowledge, we created the following prompts:
\begin{quotation}
	\noindent
	\small
	\texttt{Write a long paragraph describing the \textbf{metacognitive knowledge} presented in the following passage.}
\end{quotation}

\begin{quotation}
	\noindent
	\small
	\texttt{Write a long paragraph describing the \textbf{strategic knowledge} presented in the following passage.}
\end{quotation}

\begin{quotation}
	\noindent
	\small
	\texttt{Write a long paragraph describing the \textbf{cognitive tasks, including relevant contextual and conditional knowledge}, presented in the following passage.}
\end{quotation}

\begin{quotation}
	\noindent
	\small
	\texttt{Write a long paragraph describing the \textbf{self-knowledge} presented in the following passage.}
\end{quotation}

To get descriptions about the conventional and sitiational knowledge of the author, we created the following prompts:
\begin{quotation}
	\noindent
	\small
	\texttt{Write a long paragraph describing the \textbf{conventions} presented in the following passage.}
\end{quotation}

\begin{quotation}
	\noindent
	\small
	\texttt{Write a long paragraph describing the \textbf{situational knowledge} presented in the following passage.}
\end{quotation}

Following \citet{patel2023}, to rewrite the knowledge descriptions into lists of knowledge attributes, we designed a new rewrite prompt. Since for our approach it is important that the sentences are understandable on its own, we include an additional instruction:
\begin{quotation}
	\noindent
	\small
	\texttt{Here's a description of an author's knowledge: ...\newline \newline Rewrite this description as a long list of short sentences describing the author's knowledge where each sentence is in the format of 'The author knows X.' or 'The author understands X.' These sentences have to be short. Each sentence must be understandable on its own, without the need for context from the other sentences. Avoid unnecessary detail. Avoid examples. Avoid negations. Use only the information that is present in the description.\newline \newline Output:}
\end{quotation}

\subsection{Attribute Extraction}
\label{appendix:attribute-extracation}
We run the two-stage prompting process of our approach to extract style and knowledge attributes. We used a fixed random seed to ensure reproducibility. The prompts can be found in Appendix \ref{appendix:style-attributes} and \ref{appendix:knowledge-attributes}. For this, we randomly sampled 500 answers per group from the eleven target groups (i.e., 5500 in total) from the Stack Exchange data (see Appendix~\ref{appendix:stackexchange}). This way, we keep the total number of inference steps comparable to \citet{patel2023} who used 10k posts (10 per author), since we increase the total number of prompts to 110. In particular, for the two-stage prompting, our setup results in 1.21M inference steps, compared to 1.86M for \citet{patel2023}. 
We got 605k lists of attributes from the two-stage prompting process.

To ensure comparability across groups, we select a final shared set of style attributes $F^{(S)}$ and knowledge attributes $F^{(K)}$. For this, we apply the LLM-based filtering to keep only attributes that either belong to style or knowledge (see Appendix \ref{appendix:llm-based-filtering} for details). This resulted in about 190k unique candidate style attributes and 1.4M unique candidate knowledge attributes. Next, we clustered the attributes based on a cosine similarity threshold of 0.85 (see Appendix \ref{appendix:clustering} for details), resulting in about 31k style attribute clusters and about 470k knowledge attribute clusters. We filtered out clusters for which the representative attribute was predicted for more than seven groups. Moreover, we filtered out clusters when less then six answers were annotated with the representative attribute (see Appendix~\ref{appendix:frequency-filtering} for details). After that, for the final set of attributes $F$, we select the 86 clusters related to the targeted style prompts. From the remaining clusters, we select the representative attribute based on how often it was extracted for the Stack Exchange answers. Once an attribute is selected to be part of $F$, we do not select another attribute with a cosine similarity bigger than 0.7 to avoid too similar duplicates. We then remain with $|F| = 1250$ attributes with $F = F^{(S)} \cup F^{(K)}$ including $|F^{(S)}| = 340$ style attributes and $|F^{(K)}| = 910$ knowledge attributes.

\subsubsection{LLM-Based Filtering}
\label{appendix:llm-based-filtering}
To ensure that each extracted attribute describes either the writing style or the knowledge of an author, we apply a LLM-based filtering. In particular, we run a two-stage prompting. We used a fixed random seed to ensure reproducibility.

First, we prompt the LLM to classify each attribute into the classes \emph{knowledge}, \emph{style}, \emph{both} or \emph{none}:
\begin{quotation}
	\noindent
	\small
	\texttt{You are an expert linguist specializing in authorship analysis.\newline Your task: Given a sentence about an author, classify it into exactly one of the following four categories:\newline **knowledge**: The sentence describes what the author knows, including their skills applied.\newline **style**: The sentence describes how the author writes, including aspects such as spelling, grammar, punctuation, the vocabulary used, or the sentence and paragraph structure.\newline **both**: The sentence could describe both the author’s knowledge or their writing style.\newline **none**: The sentence does NOT describe the author's knowledge or writing style.
	}
\end{quotation}
To ensure that the LLM predicts only one of the four classes, we use guided decoding implemented in the outlines library.\footnote{\url{https://github.com/dottxt-ai/outlines}}

Second, we reclassify all attributes that were labeled with \emph{both} and reassign each one to the label \emph{knowledge} or \emph{style} that best fits.
\begin{quotation}
	\noindent
	\small
	\texttt{You are an expert linguist specializing in authorship analysis.\newline Your task: Given a sentence about an author that could refer to either the author’s knowledge or their writing style. Assign it to the single category it most accurately belongs to:\newline **knowledge**: The sentence describes what the author knows, including their skills applied.\newline **style**: The sentence describes how the author writes, including aspects such as spelling, grammar, punctuation, the vocabulary used, or the sentence and paragraph structure.}
\end{quotation}

After running the two-stage prompting, we filter out attributes labeled with \emph{none}. We retain attributes labeled with \emph{knowledge} only if they were extracted using a knowledge attribute prompt. Similarly, we retain attributes labeled with \emph{style} only if they were extracted using a style attribute prompt.

\subsubsection{Clustering}
\label{appendix:clustering}
To cluster semantically similar attributes, we perform a radius-based neighbor clustering in the embedding space based on \citet{ester1996clustering}. Each embedded attribute is treated as a node in a graph and connected to all other attributes within a radius defined by a cosine similarity threshold. Instead of forming clusters via full graph connectivity \cite{ester1996clustering}, cluster centers are identified through a greedy nearest-neighbor descent: Starting from each node, we repeatedly move to the most similar neighboring node within the radius until convergence. All attributes that converge to the same center are grouped into a cluster. This approach produces variable-sized clusters that adapt to local similarity without requiring a predefined number of clusters.

\subsubsection{Frequency-Based Filtering}
\label{appendix:frequency-filtering}
To increase the representativeness of the clusters, we filter out clusters based on two frequency-based criteria. First, we filter the representative attributes of the clusters based on a maximum number of groups for which they were extracted to avoid too general attributes, e.g., \textit{The author uses words.} We follow the filtering heuristics of \citet{patel2023} and set the maximum number to $0.6 \cdot 11 \approx 7$. In addition, to ensure that the attributes are not overly specialized, we also take into account how frequently each attribute was extracted. To determine an appropriate value, we start with 1 and gradually increase it until the resulting number of clusters is at most 10,000. This threshold is chosen to balance granularity and generalization: It is large enough to capture meaningful distinctions between attributes, but small enough to avoid overly fragmented clusters that would be sparse or noisy. Following this procedure, we arrive at a final value of six. Thus, an attribute is selected only if it was extracted for at least six answers.

\subsubsection{Ablation Study}
\label{appendix:ablation-study-attribute-extraction}

In the steering experiments, we use Qwen (7B) for both attribute extraction and explanation generation. To analyze whether this choice affects steering effectiveness of this model, we conduct an ablation study in which Llama (8B) and Ministral (8B) are used for the attribute extraction.

Executing the entire generation pipeline, including the hyperparameter tuning of the layers $I$ and the factor $\lambda$, would require a significant amount of computational resources. Therefore, we only compute the steering vectors derived from attributes extracted by Llama and Ministral (we follow the same procedure as described in \ref{appendix:attribute-extracation}). We then compare these vectors with the original steering vectors derived from Qwen-based attributes. This comparison already provides insight into the robustness of the steering, as highly similar steering vectors are expected to induce comparable hidden-state modifications and, consequently,~similar generations.

Table~\ref{tab:ablation_study_attribute_extraction_best_layers} reports the cosine similarity between steering vectors obtained from Llama- or Ministral-based attributes and the original Qwen-based steering vectors. Across all evaluated models, the steering directions remain highly aligned, with cosine similarities consistently above 0.8 and reaching up to 0.96. Notably, for Qwen, the steering vectors derived from attributes extracted by Llama and Ministral remain highly similar to the original steering vectors based on the attributes extracted by Qwen (0.9 and 0.85). This suggests that the strong steering performance observed when using Qwen for generating the explanations (see Table \ref{tab:steering_eval_science_questions} and \ref{tab:human_study_both}) is not a consequence of Qwen extracting the attributes itself.

Overall, the results indicates that our method is robust to variations in attribute extraction and does not critically depend on using Qwen. Instead, the steering directions appears to be highly similar across different LLMs.

\begin{table*}[t]
	\small
	\centering
	\renewcommand{\arraystretch}{0.95}
	\setlength{\tabcolsep}{5pt}
	\begin{tabular}{lllrr}
		\toprule
		\textbf{LLM} & \textbf{Steering Vectors} & \textbf{Layers $I$} &
		\textbf{Llama-based} & \textbf{Ministral-based} \\
		\midrule
		\multirow{1}{*}{\rotatebox{0}{Qwen 7B}}
		& \textbf{Our approach (5)} & [10,11,12] & $0.90 \pm 0.04$ & $0.85 \pm 0.04$\\
		\midrule
		\multirow{1}{*}{\rotatebox{0}{Llama 8B}}
		& \textbf{Our approach (5)} & [17,18,19] & $0.86 \pm 0.05$& $0.81 \pm 0.05$\\
		\midrule
		\multirow{1}{*}{\rotatebox{0}{Ministral 8B}}
		& \textbf{Our approach (5)} & [5,6,7] & $0.96 \pm 0.01$ & $0.93 \pm 0.02$\\
		\bottomrule
	\end{tabular}
	\caption{
		Cosine similarity of the steering vectors based on the attributes extracted by Llama (\textit{Llama-based}) or Ministral (\textit{Ministral-based}) and the original steering vectors based on the attributes extracted by Qwen. The values represent the cosine similarity averaged across all steered layers $I$ and all 11 target groups, as well as the standard deviation.
	}
	\label{tab:ablation_study_attribute_extraction_best_layers}
\end{table*}

\subsection{Most Important Attributes}
\label{appendix:important-attributes}
To identify the $m$ most important style and the $m$ most important knowledge attributes per group~$g$, we compute the point-biserial correlation \cite{lev1949PointBiserialCorrelation} for all attribute vectors $\mathbf{v}_g \in \{0,1\}^{1250}$ created for each of the 500 answers from Stack Exchange during the two-stage prompting process. We also considered to identify the most important attributes based on the attributes vectors predicted by the model described in Appendix~\ref{appendix:group-score} for the 6000 answers per group . However, the most important attributes identified based on these attribute vectors seem to be less representative than those based on the 500 answers. One reason for this could be that predicting the attribute vectors for the 6000 answers using the model is impacted by the inherent generalization error. While this does not have a big effect when considering the whole attribute vector, the attribute-specific quality is lower. This is why we identify the most important attributes based on the 500 answers per group. Table \ref{tab:style-attributes} shows the most important style attributes for $m=5$ per group ordered by the correlation values. Table \ref{tab:knowledge-attributes} lists the top five most important knowledge attributes for for each group, also ranked by their correlation values.

\begin{table*}[ht]
	\centering
	\small
	\begin{tabular}{l|p{9cm}|c}
		\toprule
		\textbf{Group} & \textbf{Style Attribute} & \textbf{Correlation} \\
		\midrule
		
		\multirow{5}{*}{Pilots}
		& The author uses technical aviation terms. & 0.5418 \\
		& The author uses regulatory language. & 0.2098 \\
		& The author refers to pilots without using gender-specific terms. & 0.1409 \\
		& The author emphasizes coordination. & 0.0811 \\
		& The author uses altitude descriptions. & 0.0801 \\
		\midrule
		
		\multirow{5}{*}{Biologists}
		& The author uses medical terminology. & 0.3137 \\
		& The author uses mutations. & 0.1532 \\
		& The author uses technical terms from genetic and molecular biology. & 0.1350 \\
		& The author uses physiological descriptions. & 0.1201 \\
		& The author conveys complex scientific concepts clearly. & 0.1084 \\
		\midrule
		
		\multirow{5}{*}{Chemists}
		& The author uses chemical notation. & 0.5921 \\
		& The author uses the \texttt{\textbackslash ce\{\}} command. & 0.3149 \\
		& The author focuses on reactions. & 0.1994 \\
		& The author conveys complex scientific concepts clearly. & 0.1198 \\
		& The author uses degrees Celsius. & 0.1191 \\
		\midrule
		
		\multirow{5}{*}{Computer scientists}
		& The author uses formal language theory. & 0.2372 \\
		& The author uses O notation. & 0.1859 \\
		& The author uses inequalities. & 0.1674 \\
		& The author uses descriptions of algorithms. & 0.1620 \\
		& The author uses backslashes before certain symbols. & 0.1426 \\
		\midrule
		
		\multirow{5}{*}{Electrical engineers}
		& The author uses specific electrical engineering terminology. & 0.1480 \\
		& The author uses synchronization. & 0.0665 \\
		& The author uses backslashes before certain symbols. & 0.0575 \\
		& The author uses commas for technical descriptions. & 0.0483 \\
		& The author logically progresses from definitions to implications. & 0.0470 \\
		\midrule
		
		\multirow{5}{*}{Game developers}
		& The author includes code snippets. & 0.3132 \\
		& The author uses terms related to video games. & 0.2732 \\
		& The author uses function names. & 0.1420 \\
		& The author uses \texttt{//} for comments. & 0.1208 \\
		& The author uses C\# syntax. & 0.1208 \\
		\midrule
		
		\multirow{5}{*}{Historians}
		& The author adheres to historical writing conventions. & 0.2707 \\
		& The author uses World War II as a backdrop. & 0.1479 \\
		& The author uses words related to time. & 0.1240 \\
		& The author uses political terminology. & 0.1218 \\
		& The author emphasizes event significance. & 0.1133 \\
		\midrule
		
		\multirow{5}{*}{Philosophers}
		& The author grapples with existential questions. & 0.2091 \\
		& The author uses religious perspectives. & 0.1672 \\
		& The author uses moral considerations. & 0.1667 \\
		& The author approaches argumentation deliberately. & 0.1038 \\
		& The author uses tautologies. & 0.0970 \\
		\midrule
		
		\multirow{5}{*}{Physicists}
		& The author writes in a style common to physics literature. & 0.2216 \\
		& The author formats mathematical expressions correctly. & 0.1528 \\
		& The author conveys complex scientific concepts clearly. & 0.1501 \\
		& The author uses ket notation. & 0.1201 \\
		& The author uses constants. & 0.1030 \\
		\midrule
		
		\multirow{5}{*}{Politicians}
		& The author uses political terminology. & 0.5493 \\
		& The author uses sarcasm. & 0.1480 \\
		& The author underscores gravity. & 0.1303 \\
		& The author uses media representation. & 0.1207 \\
		& The author uses language related to Russia's actions. & 0.1045 \\
		\midrule
		
		\multirow{5}{*}{Software engineers}
		& The author emphasizes maintainability. & 0.1422 \\
		& The author uses colloquial expressions. & 0.1269 \\
		& The author emphasizes personal growth. & 0.1069 \\
		& The author uses words related to work. & 0.0966 \\
		& The author uses terms related to testing. & 0.0963 \\
		\midrule
		
	\end{tabular}
	\caption{Top five most important style attributes (by point-biserial correlation) for each group based on the attribute vectors $\mathbf{v}_g \in \{0,1\}^{1250}$ created for the 500 answers per group from Stack Exchange. \label{tab:style-attributes}}
\end{table*}

\begin{table*}[ht]
	\centering
	\small
	\begin{tabular}{l|p{10cm}|c}
		\toprule
		\textbf{Group} & \textbf{Knowledge Attribute} & \textbf{Correlation} \\
		\midrule
		
		\multirow{5}{*}{Pilots}
		& The author understands aerodynamics. & 0.5925 \\
		& The author knows about the relationship between airline operations and safety regulations. & 0.5365 \\
		& The author understands pilots must be adaptable. & 0.3668 \\
		& The author understands fuel efficiency. & 0.3078 \\
		& The author knows modern aircraft systems are sophisticated. & 0.3029 \\
		\midrule
		
		\multirow{5}{*}{Biologists}
		& The author understands evolutionary biology. & 0.4238 \\
		& The author understands metabolic pathways. & 0.2418 \\
		& The author understands environmental and genetic factors interact. & 0.2170 \\
		& The author understands the immune system plays a role. & 0.2128 \\
		& The author understands mutation introduces genetic variation. & 0.2087 \\
		\midrule
		
		\multirow{5}{*}{Chemists}
		& The author understands chemical bonding. & 0.4000 \\
		& The author knows redox reactions involve electron transfer. & 0.2661 \\
		& The author understands the importance of reaction conditions. & 0.2292 \\
		& The author understands pH measures hydrogen ion concentration. & 0.2211 \\
		& The author knows the ideal gas law. & 0.2103 \\
		\midrule
		
		\multirow{5}{*}{Computer scientists}
		& The author knows computational complexity theory. & 0.4334 \\
		& The author understands set theory. & 0.2369 \\
		& The author understands input size affects algorithm performance. & 0.2084 \\
		& The author knows developing efficient algorithms is necessary. & 0.1813 \\
		& The author knows deterministic finite automata (DFAs). & 0.1708 \\
		\midrule
		
		\multirow{5}{*}{Electrical engineers}
		& The author understands voltage. & 0.4770 \\
		& The author understands the importance of component selection. & 0.3502 \\
		& The author knows microcontrollers. & 0.3283 \\
		& The author understands the importance of signal fidelity. & 0.2740 \\
		& The author understands the importance of power supply efficiency. & 0.2381 \\
		\midrule
		
		\multirow{5}{*}{Game developers}
		& The author understands game development. & 0.6761 \\
		& The author understands rendering techniques. & 0.4227 \\
		& The author understands this approach enhances the gaming experience. & 0.2769 \\
		& The author understands collision detection. & 0.2629 \\
		& The author knows the importance of maintaining a consistent frame rate. & 0.2375 \\
		\midrule
		
		\multirow{5}{*}{Historians}
		& The author understands military strategy. & 0.4075 \\
		& The author knows historical analysis is important. & 0.3329 \\
		& The author understands political dynamics. & 0.2744 \\
		& The author understands the distinction between primary and secondary sources. & 0.2614 \\
		& The author understands archaeological evidence. & 0.2556 \\
		\midrule
		
		\multirow{5}{*}{Philosophers}
		& The author understands consciousness. & 0.2732 \\
		& The author knows belief systems exist. & 0.2527 \\
		& The author knows Immanuel Kant's philosophy. & 0.2457 \\
		& The author knows logical fallacies exist. & 0.2148 \\
		& The author understands propositions can be true or false. & 0.2128 \\
		\midrule
		
		\multirow{5}{*}{Physicists}
		& The author knows quantum mechanics. & 0.3935 \\
		& The author knows special relativity. & 0.2672 \\
		& The author understands momentum conservation. & 0.2524 \\
		& The author understands magnetic fields. & 0.2497 \\
		& The author understands wave propagation. & 0.2478 \\
		\midrule
		
		\multirow{5}{*}{Politicians}
		& The author understands political dynamics. & 0.5887 \\
		& The author knows constitutional law. & 0.3552 \\
		& The author understands media influence. & 0.2988 \\
		& The author understands the role of international cooperation. & 0.2429 \\
		& The author knows about Russia. & 0.2212 \\
		\midrule
		
		\multirow{5}{*}{Software engineers}
		& The author knows the importance of clarity in code. & 0.3007 \\
		& The author understands project management. & 0.2552 \\
		& The author knows metacognitive knowledge applies to software development. & 0.2323 \\
		& The author knows the importance of unit testing. & 0.2211 \\
		& The author knows version control systems. & 0.2187 \\
		\midrule
		
	\end{tabular}
	\caption{Top five most important knowledge attributes (by point-biserial correlation) for each group based on the attribute vectors $\mathbf{v}_g \in \{0,1\}^{1250}$ created for the 500 answers per group from Stack Exchange. \label{tab:knowledge-attributes}}
\end{table*}

\section{Group-Specific Explanation Generation}
\label{appendix:explanation-gen}

\subsection{GPU usage}
We conducted all experiments on a shared cluster equipped with 16 H200 and 8 A100, with a total compute cost of approximately 200 GPU hours.

\subsection{Prompting Baseline}
\label{appendix:prompting-baseline}
In the experiments, we compare our approach against a prompting baseline. We investigate two different system prompt settings. Since the Stack Exchange explanations from our dataset contain around 168 words on average, we instructed the LLM to limit its responses to 200 words.

The first system prompt includes only the target group (e.g., game developers):
\begin{quotation}
	\noindent
	\raggedright
	\small
	\texttt{You are a helpful assistant specialized in answering explanatory questions asked by \textbf{game developers}. Explain in a way that \textbf{game developers} can effectively understand the explanations. Your explanations have to be shorter than 200 words.}
\end{quotation}
\noindent
The second prompt additionally includes the $m$ most important style attributes and the $m$ most important knowledge attributes for that group. We rewrite the extracted attributes (see Section~\ref{sec:approachAttributeVector}) so that it fits into the prompt (e.g., \textit{The author includes code snippets.} $\rightarrow$ \textit{They include code snippets.}): 

\begin{quotation}
	\noindent
	\raggedright
	\small
	\texttt{Take into account that \textbf{game developers} have the following explanatory style:\newline They include code snippets. \newline... \newline \newline Take into account that \textbf{game developers} usually have the following knowledge: \newline They understand game development. \newline ... \newline \newline You are a helpful assistant specialized in answering explanatory questions asked by \textbf{game developers}. Explain in a way that \textbf{game developers} can effectively understand the explanations. Your explanations have to be shorter than 200 words.}
\end{quotation}

\subsection{Specificity Metric}
\label{appendix:group-score}
To measure the group specificity of a generated explanation, we train a classifier $c: [0,1]^{1250} \rightarrow \mathbb{R}^{11}$ that predicts a value for each target group based on the attribute vector created for that explanation. For this, we follow \citet{patel2023} by first training a model (SFAM) based on \texttt{deberta-v3-large} that takes an input text and predicts a probability value for each attribute $f_j \in F$ how likely the attribute is present in the text. To train and evaluate this model, we split the Stack Exchange dataset into 400 train samples, 50 validation samples and 50 test samples per group. This model is then trained on the attribute vectors $\mathbf{v} \in \{0,1\}^{1250}$ created for the 400 answers per group~$g$ during the two-stage prompting process. The test accuracy of that model is $0.876$. With this model, we extend our Stack Exchange dataset by constructing the attribute vectors $\mathbf{v} \in [0,1]^{1250}$ for all 6000 answers per group. We again split the data into 5500 train samples, 300 validation samples and 200 test samples per group. Based on these data, we train another model (LISA) based on \texttt{deberta-v3-large} that directly maps an input text to an attribute vector $\mathbf{v} \in [0,1]^{1250}$. It enables the fast computation of an attribute vector for a generated explanation. This model achieves a mean squared error of $0.0437$ on the test set. Finally, we train the classifier on the 6000 attribute vectors per group (5500/300/200). The classifier achieves a test accuracy of $0.852$ on the test set. Given a generated explanation tailored to a specific group and the corresponding attribute vector $\mathbf{v}$, the $specificity \in [0, 1]$ is obtained by applying the softmax to the classifier output and selecting the component corresponding to the group. For the test set, the classifier achieves an average $specificity$ score of $0.801$.

The group specificity serves also as the validation metric to find the best values for the hyperparameters $I$ and $\lambda$ on the validation set. However, since activation-based steering leads the LLM to generate endless repetitive texts when $\lambda$ is chosen too high \cite{konen2024steeringstyle}, we additionally include a binary repetition penalty $r$ and length penalty $l$. In addition, we check that the generated explanations contain only Latin-1 characters since we want to generate explanations in English. Since we prioritize fluent explanations, we weighted the repetition penalty twice as much as the length penalty for obtaining $specificity_{val} \in [-3,1]$: 
\begin{equation}
	specificity_{val} \coloneqq \textrm{softmax}(c(\mathbf{v}))_g  - 2 \cdot r - l
\end{equation}

Since the human study revealed that the steering success of the explanations generated by our approach was nearly maximum $(4.17/5)$, we conclude that the quality of the classifier is sufficient for hyperparameter tuning. Therefore, we emphasize its real-world usefulness in general.

\subsection{Factuality Metric}
\label{appendix:factscore}
To measure the factuality of a generated explanations, we use the \textsc{FActScore} metric \cite{min2023factscore} which computes the probability of the set of atomic facts extracted from the explanation that are supported by a reliable knowledge source.

To foster reproducibility, we utilize the open-source implementation \cite{lage2025openfactscore}. As \citet{lage2025openfactscore} suggest, we selected \texttt{OLMo-2-1124-7B-SFT} for the atomic fact generation, and \texttt{gemma-3-4b-it} for the atomic fact validation according to Wikipedia (as of: April~1,~2023). Since activation-based steering leads the LLM to generate nonsense texts when $\lambda$ is chosen too high \cite{konen2024steeringstyle}, \texttt{OLMo-2-1124-7B-SFT} fails to generate atomic facts for such explanations in our experiment. We decided to set the factuality score to zero in such cases since the implementation of \citet{lage2025openfactscore} does not handle this.

\subsection{Harmonic Mean Metric}
\label{appendix:hmean}
We additionally compute the harmonic mean to capture the balance of the evaluation criteria. Since the harmonic mean is calculated based on the mean values, we cannot report significance results for the harmonic mean.

For the automatic evaluation, we compute the harmonic mean based on the mean values for specificity (s) and factuality (f):
\begin{equation}
	\text{HMean} \coloneqq \frac{2 \cdot s \cdot f}{s + f}
\end{equation}

For the manual evaluation, we compute the harmonic mean based on the mean values for steering success (s), helpfulness (h) and plausibility (p):
\begin{equation}
	\text{HMean} \coloneqq 
	\frac{3 \cdot s \cdot h \cdot p}{s \cdot h + s \cdot p + h \cdot p}
\end{equation}

\subsection{Results for ELI5 Questions}
\label{appendix:results-eli5}
Table \ref{tab:steering_eval_eli5} shows the mean results of our main experiments for the 600 ELI5 questions where we evaluate our approach based on the $m=5$ most important attributes against two baselines using different model families. We did not compute the $factuality$ since the \textsc{FActScore} metric \cite{min2023factscore} requires specifying a topic that maps to a Wikipedia title but not all ELI5 questions can be clearly mapped to a Wikipedia title. 

\begin{table*}[t]
	\small
	\centering
	\begin{tabular}{lllllr}
		\toprule
		\textbf{LLM} & \textbf{Prompt Components} & \textbf{Steering Vectors} &
		\textbf{Factor$\lambda$} & \textbf{Layers $I$} &
		\textbf{Specificity $\uparrow$} \\
		\midrule
		\multirow{6}{*}{\rotatebox{90}{Qwen (7B)}}
		& group & - & - & - & 0.139\\
		& group + \textbf{attributes (5)} & - & - & - & 0.239 \\
		& group & \citet{konen2024steeringstyle} & 1.75 & [17,18,19] & 0.747 \\
		& group + \textbf{attributes (5)} & \citet{konen2024steeringstyle} & 1.50 & [18,19,20] & \textbf{0.827} \\
		& group & \textbf{Our approach (5)} & 2.00 & [10,11,12] & 0.382\\
		& group + \textbf{attributes (5)} & \textbf{Our approach (5)} & 1.50 & [10,11,12] & 0.457 \\
		
		\midrule
		\multirow{6}{*}{\rotatebox{90}{Llama (8B)}}
		& group & - & - & - & 0.214 \\
		& group + \textbf{attributes (5)} & - & - & - & 0.425 \\
		& group & \citet{konen2024steeringstyle} & 1.25 & [15,16,17] & 0.563 \\
		& group + \textbf{attributes (5)} & \citet{konen2024steeringstyle} & 1.00 & [17,18,19] & \textbf{0.701} \\
		& group & \textbf{Our approach (5)} & 0.75 & [17,18,19] & 0.316 \\
		& group + \textbf{attributes (5)} & \textbf{Our approach (5)} & 0.75 & [17,18,19] & 0.511 \\
		
		\midrule
		\multirow{6}{*}{\rotatebox{90}{Ministral (8B)}}
		& group & - & - & - & 0.139 \\
		& group + \textbf{attributes (5)} & - & - & - &  0.266\\
		& group & \citet{konen2024steeringstyle} & 1.00 & [20,21,22] & 0.366 \\
		& group + \textbf{attributes (5)} & \citet{konen2024steeringstyle} & 1.00 & [21,22,23] & \textbf{0.584} \\
		& group & \textbf{Our approach (5)} & 1.00 & [5,6,7] & 0.171 \\
		& group + \textbf{attributes (5)} & \textbf{Our approach (5)} & 1.00 & [5,6,7] & 0.336 \\
		
		\bottomrule 
	\end{tabular}
	\caption{Main automatic evaluation results: \emph{Specificity} of all tested \emph{LLMs}, \emph{prompt components}, and \emph{steering vectors} on the ELI5 test questions, along with steering \emph{factor} $\lambda$ and steered \emph{layers} $I$. Bold components/vectors are part of our approach. Best values per LLM bold.}
	\label{tab:steering_eval_eli5}
\end{table*}

\subsection{Ablation Results}
\label{appendix:results-ablations}
\begin{table*}[t]
	\small
	\centering
	\renewcommand{\arraystretch}{0.95}
	\setlength{\tabcolsep}{5pt}
	\begin{tabular}{lllrlrrr}
			\toprule
		\textbf{LLM} & \textbf{Prompt Components} & \textbf{Steering Vectors} &
		\textbf{Factor} $\mathbf{\lambda}$ & \textbf{Layers $I$} &
		\textbf{Specificity $\uparrow$} & \textbf{Factuality $\uparrow$} & \textbf{HMean $\uparrow$} \\
		\midrule
		\multirow{4}{*}{\rotatebox{90}{Qwen 7B}}
		& Group & \textbf{Our approach (3)} & 1.75 & [10,11,12] & 0.356 & 0.279 & 0.313 \\
		& Group \textbf{+ attributes (3)} & \textbf{Our approach (3)} & 1.50 & [10,11,12] & 0.433 & 0.230 & 0.300\\		
		& Group & \textbf{Our approach (10)} & 2.00 & [11,12,13] & 0.366 & \textbf{0.281} & \textbf{0.318} \\
		& Group \textbf{+ attributes (10)} & \textbf{Our approach (10)} & 1.50 & [10,11,12] & \textbf{0.439} & 0.228 & 0.300 \\
		\midrule
		\multirow{4}{*}{\rotatebox{90}{Qwen {32B}}}
		& Group & - & - & - & 0.144 & 0.328 & 0.200 \\
		& Group \textbf{+ attributes (10)} & - & - & - & 0.229 & 0.289 & 0.255 \\
		& Group & \textbf{Our approach (10)} & 1.00 & [41,42,43] & 0.187 & \textbf{0.370} & 0.249 \\
		& Group \textbf{+ attributes (10)} & \textbf{Our approach (10)} &  2.00   & [38,39,40] & \textbf{0.374} & 0.315 & \textbf{0.342} \\
		\bottomrule
	\end{tabular}
	\caption{
	Automatic ablation results, comparing \emph{Qwen 7B} and \emph{32B} with 3 and 10 attributes respectively on the science questions. Best values per LLM bold. See caption of Table~\ref{tab:steering_eval_science_questions} for further explanations.
	}
	\label{tab:steering_eval_ablations_science_questions}
\end{table*}

\begin{table*}[t]
	\small
	\centering
	\begin{tabular}{lllllr}
		\toprule
		\textbf{LLM} & \textbf{Prompt Components} & \textbf{Steering Vectors} &
		\textbf{Factor $\lambda$} & \textbf{Layers $I$} &
		\textbf{Specificity $\uparrow$} \\
		\midrule
		\multirow{4}{*}{\rotatebox{90}{Qwen 7B}}
		& group & \textbf{Our approach (3)} & 1.75 & [10,11,12] & 0.365 \\
		& group + \textbf{attributes (3)} & \textbf{Our approach (3)} & 1.50 & [10,11,12] & 0.445\\
		& group & \textbf{Our approach (10)} & 2.00 & [11,12,13] & 0.379 \\
		& group + \textbf{attributes (10)} & \textbf{Our approach (10)} & 1.50 & [10,11,12] & \textbf{0.458} \\
		
		\midrule
		\multirow{4}{*}{\rotatebox{90}{Qwen 32B}}
		& group & - & - & - & 0.157 \\
		& group + \textbf{attributes (10)} & - & - & - & 0.256 \\
		& group & \textbf{Our approach (10)} & 1.00 & [41,42,43] & 0.213 \\
		& group + \textbf{attributes (10)} & \textbf{Our approach (10)} & 2.00 & [38,39,40] & \textbf{0.480} \\
		
		\bottomrule
	\end{tabular}
	\caption{Automatic ablation results, comparing \emph{Qwen 7B} and \emph{32B} with 3 and 10 attributes respectively on the ELI5 test questions. Best values per LLM bold. See caption of Table~\ref{tab:steering_eval_eli5} for further explanations.}
	\label{tab:steering_eval_ablations_eli5}
\end{table*}

As described in Appendix \ref{appendix:important-attributes}, we identify the top $m$ most important style attributes and the $m$ most important knowledge attributes per group. To assess the impact of $m$ on the steering effectiveness, we consider $m \in \{3, 5, 10\}$ for the group-specific explanation generation. Table~\ref{tab:steering_eval_ablations_science_questions} shows the results of our approach based on the Qwen model when using the $m=3$ and $m=10$ most important attributes for the 97 science questions. To further examine how model capacity influences steering effectiveness, we additionally evaluate a larger variant from the Qwen family (\texttt{Qwen2.5-32B-Instruct}).
Table \ref{tab:steering_eval_ablations_eli5} shows the mean results for the ELI5 questions.

Overall, the results denote that there are only marginal performance differences between $m=3$ and $m=10$ across all dimensions indicating we found a sweet spot with $m=5$. The results of the science question shown in Table~\ref{tab:steering_eval_ablations_science_questions} indicate that increasing the model size increases the factuality but decreases the specificity of the explanations. Based on the results shown in Table \ref{tab:steering_eval_science_questions} and ~\ref{tab:steering_eval_ablations_science_questions}, using Qwen 32B with the second system prompt setting achieves the best balance between specificity and factuality ($\textrm{HMean}=0.342$). For the ELI5 question shown in Table~\ref{tab:steering_eval_ablations_eli5}, increasing the model size resulted in increased specificity. These findings suggest that using larger models can help achieving a better balance between specificity and factuality.

\subsection{Results per Group}
\label{appendix:specificty-per-group}
Figure \ref{fig:group-score-per-group-science-questions} and \ref{fig:group-score-per-group-eli5-questions} illustrate the steering effectiveness in terms of specificity per group of the prompting baseline, the standard activation-based steering baseline \cite{konen2024steeringstyle} and our approach. In Figure \ref{fig:group-score-per-group-science-questions}, only the explanations for the 97 science questions are evaluated. Figure \ref{fig:group-score-per-group-eli5-questions} shows the results for the 600 ELI5 test questions. As a result, the approach of \citet{konen2024steeringstyle} has the highest specificity for nearly all target groups across both question types. The prompting baseline the lowest and our approach is in the middle. The highest specificity was achieved by our approach for philosophers. The approach of \citet{konen2024steeringstyle} additionally achieves high specificity values for software engineers and game developers, for whom the performance of the prompting baseline collapses to nearly zero for the science questions.

\bsfigure[width=0.49\textwidth]{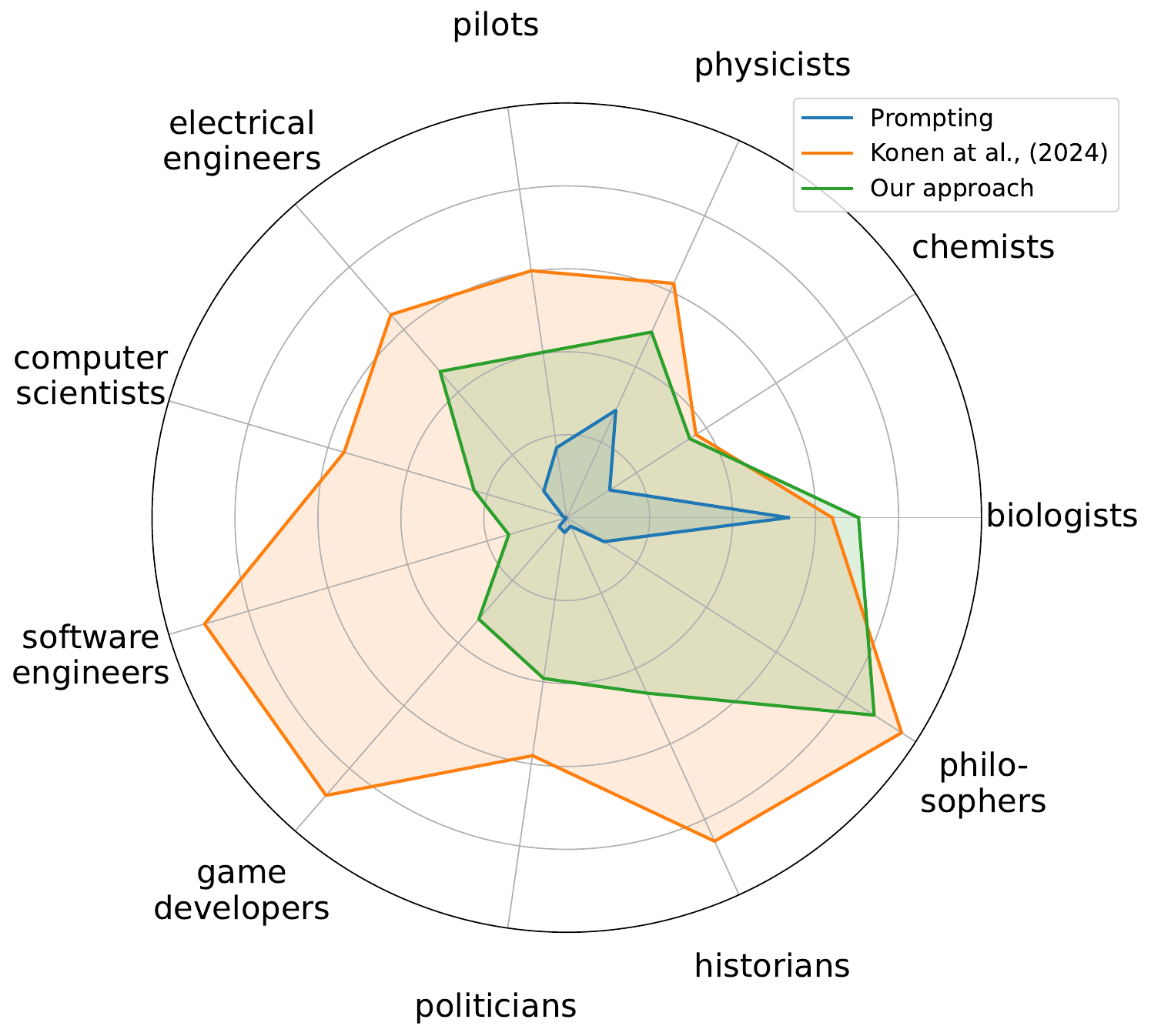}{Per-group steering effectiveness based on the mean $specificity$ scores for the 97 science questions. The blue area shows the prompting and the orange area shows the activation-based steering baseline \cite{konen2024steeringstyle}, each using only the group in the prompt. The green area shows our approach, which additionally uses the five most important attributes in the prompt. \label{fig:group-score-per-group-science-questions}}

\bsfigure[width=0.49\textwidth]{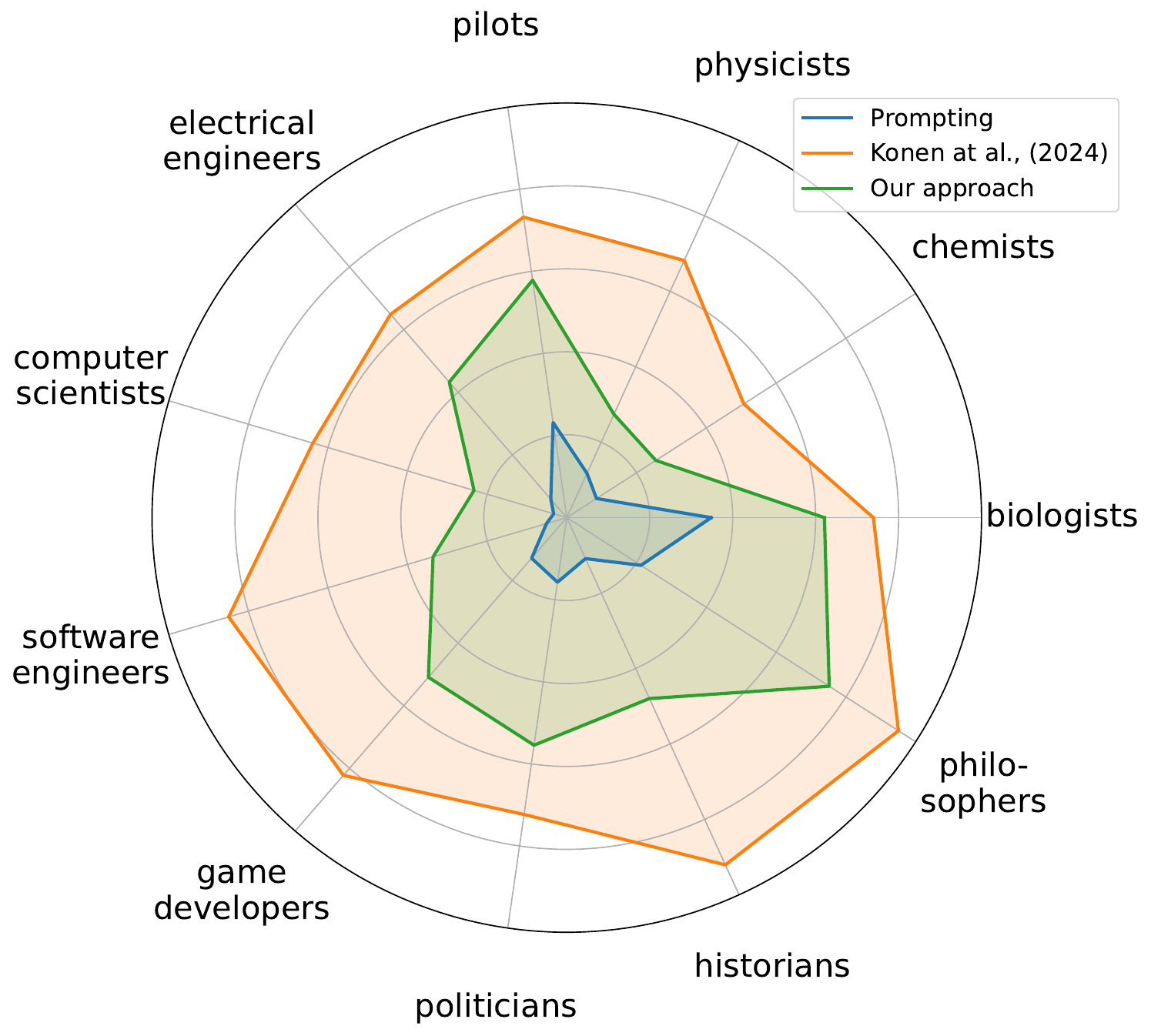}{Per-group steering effectiveness based on the mean $specificity$ scores for the 600 ELI5 questions. The blue area shows the prompting and the orange area shows the activation-based steering baseline \cite{konen2024steeringstyle}, each using only the group in the prompt. The green area shows our approach, which additionally uses the five most important attributes in the prompt. \label{fig:group-score-per-group-eli5-questions}}

\subsection{Human Study}
\label{appendix:human-study}
To manually assess the steering effectiveness of the generated explanations of the Qwen (7B) model, we conducted a user study on the online freelancing platform Upwork\footnote{Upwork, \url{https://www.upwork.com}}. We hire nine participants consisting of three experts representing the target groups: biologists, game developers, and philosophers. We selected these target groups since they are the best performing groups across all approaches of the three clusters (see Figure \ref{fig:profile-correlations} and \ref{fig:hmean-per-group-science-questions}). While a study with nine experts \emph{might} be limited, we would like to point out that having three experts for each of three groups is more than is common in many NLP generation papers, in our experience. We define an expert as someone with formal education (e.g., a degree in the respective field) or professional experience in the field. In addition, we filter for people with a job success rate of at least 90\% to ensure high quality results.

We sample 25 questions from the 97 science questions and 25 questions from the 600 ELI5 test questions. Thus, the task of the study consists of 50  randomly sampled questions with three explanations for each question. The first explanation was generated by the prompting baseline that only includes the target group in the prompt. The second explanation was generated by the approach of \citet{konen2024steeringstyle}, only based on the prompt including the target group. The third explanation was generated by our approach based on the five most important style and knowledge attributes. We randomly shuffle the order of the three explanations for each question to avoid a learning effect. The participants were instructed to assume that a person of their target group (e.g., a game developer) asked the question and is honestly interested in the explanations. This was important to avoid a misunderstanding when the question is in the domain of the target group, e.g., ``How does DNA work?'' for biologists, since normally an expert in biology would already know the answer. The task was to rate the explanations based on the following three criteria:
\begin{enumerate}
	\item How well is each explanation tailored to <target group>?
	\item How helpful is each explanation for <target group> to understand the topic?
	\item Does each explanation seem plausible to you?
\end{enumerate}
We assess the three criteria on a 5-point Likert scale from \emph{not at all} (1) to \emph{fully} (5) using Google forms. We estimate that it takes about seven minutes to read the explanations and to answer our questions, which results in about six hours of work in total. We paid a fixed-price of $\$100$ which is a hourly wage of about $\$16$ which we assume to be an adequate pay anywhere on earth.

In the following, the job description for biologists posted on Upwork is shown:
\begin{quotation}
	\noindent
	\raggedright
	\small
	\texttt{In this study, our goal is to investigate how to best explain a topic to people with different backgrounds and abilities. \newline The task consists of 50 questions with three explanations for each question. You will be asked to judge how well the explanations are tailored to biologists. You will also be asked to rate whether the explanations are helpful for understanding the topic and whether the content is plausible. \newline We estimate that it will take about 7 minutes to read the three explanations and to answer our questions, which results in about 6 hours of work in total. The deadline for submission is 31.12.2025, 11:59 PM (CET). \newline Requirements: \newline- You must have a formal education (e.g., a degree in biology) or professional experience in the field of biology.
	\newline - You must be fluent in English, with the ability to understand scientific explanations.}
\end{quotation}

In the following, the instruction for the participants (in this case biologists) are shown:
\begin{quotation}
	\noindent
	\raggedright
	\small
	\texttt{Hi xxx,
		\newline Thank you for your quick reply! :)
		\newline As mentioned in the job description, the goal is to compare different explanations and evaluate how well they are tailored to biologists. A fully tailored explanation considers the background knowledge and experience of biologists.
		\newline You will be given 50 questions (e.g., Why is the sky blue?) and for each of the questions three explanations (A/B/C). Assume that a biologist asked these questions and is honestly interested in the explanations. Since you have experience in biology, your task is to rate the explanations based on the following three criteria:
		\newline (1) How well is each explanation tailored to biologists?
		\newline (2) How helpful is each explanation for biologists to understand the topic?
		\newline (3) Does each explanation seem plausible to you?
		\newline Rate the explanations on a 5-point scale, from "not at all" to "fully".
		\newline IMPORTANT: Evaluate each criterion independently. For example, an explanation may be fully tailored by making use of good analogies, yet the topic may still be difficult to understand. In that case, rate the first question as "fully" and the second question as "not at all," not low for both.
		\newline You will get a link to a test Google form, which includes only one question. This is to ensure that you can open, navigate, and submit the Google form without any issues. Once you have successfully submitted the test form, I will send you a link to the final Google form with all 50 questions.
		\newline Note: If you want to save your progress and reload it at a later point, you will need to log in with a Google account. We would recommend conducting the study on a laptop or tablet, not on a smartphone.
		\newline If you have any questions, please don’t hesitate to reach out! Otherwise, let me know if you are ready to start the study. :)
		\newline Test Link: <Google forms link>
		\newline Participant ID: <id>
		\newline Best,
		\newline xxx}
\end{quotation}

Table \ref{tab:human_study_eli5} shows the mean results in terms of steering success (first question), helpfulness (second questions) and plausibility (third question) assessed for the generated explanations of the ELI5 test questions. Significance difference it determined with the Wilcoxon signed-rank test. In addition, the inter-annotator agreement (IAA) based on the Krippendorff’s alpha is shown. Table \ref{tab:human_study_rooein} lists the mean results for the science questions. Our approach achieves the highest steering success across both question types by a significant margin. While the approach of \citet{konen2024steeringstyle} has the lowest values for helpfulness and plausibility, the prompting baseline performs best for these two dimensions. However, there is no significant difference in the helpfulness values ($4.01$ vs. $3.90$) of the prompting baseline and our approach for the science questions as shown in Table \ref{tab:human_study_rooein} in contrast to the ELI5 questions (see Table \ref{tab:human_study_eli5}). We speculate that the complexity of the science questions enables more benefit of group-specific explanations as generated by our approach, while the rather general topics of the ELI5 questions could already be well-understood with generic explanations as generated by the prompting baseline.

While the IAA is low to moderate for all evaluation criteria, this is expected for subjective tasks, since the goal is not to achieve a high IAA, as it is in annotation. Rather, the IAA serves as a means to quantify the amount of subjectivity in the average of various opinions. Our manual inspection of the evaluators’ results shows that they all worked reliably.

\begin{table*}[t]
	\small
	\centering
	\begin{tabular}{llrrrrrr}
		\toprule
		&  &
		\multicolumn{2}{c}{\textbf{Steering Success}} &
		\multicolumn{2}{c}{\textbf{Plausibility}} &
		\multicolumn{2}{c}{\textbf{Helpfulness}} \\
		\cmidrule(lr){3-4}
		\cmidrule(lr){5-6}
		\cmidrule(lr){7-8}
		\textbf{Prompt Components} & \textbf{Steering Vectors} & \textbf{Mean{$\;\pm$std}} & \textbf{IAA} & \textbf{Mean{$\;\pm$std}} & \textbf{IAA} & \textbf{Mean{$\;\pm$std}} & \textbf{IAA} \\
		\midrule
		group & - & \textsuperscript{$\dagger\!$}2.95$\;\pm1.28$ & 0.139 & \textsuperscript{$\dagger\!$}4.55$\;\pm0.83$ & 0.352 & \textsuperscript{$\dagger\!$}4.04$\;\pm1.09$ & 0.464 \\
		group & \citet{konen2024steeringstyle} & \textsuperscript{$\ddagger\!$}3.39$\;\pm1.50$ & 0.431 & \textsuperscript{$\ddagger\!$}2.65$\;\pm1.43$ & 0.254 & \textsuperscript{$\ddagger\!$}2.68$\;\pm1.39$ & 0.322 \\
		\textbf{group + attributes (5)} & \textbf{Our approach (5)} & \textsuperscript{$\dagger \ddagger\!$}3.99$\;\pm1.13$ & 0.313 & \textsuperscript{$\dagger \ddagger\!$}3.62$\;\pm1.28$ & 0.161 & \textsuperscript{$\dagger \ddagger\!$}3.39$\;\pm1.25$ & 0.178 \\
		\bottomrule
	\end{tabular}
	\caption{Manual evaluation results based on 25 randomly sampled ELI5 test questions: We evaluate \textit{steering success} in terms of how well the explanations are tailored to the target group, the \textit{plausibility} of the explanations and, how \textit{helpful} the explanations are to understand the topic generated by Qwen 7B for the two baselines and our approach. Significant difference to the prompting baseline are marked with \textsuperscript{$\dagger$} and significant difference to  \citet{konen2024steeringstyle} are marked with~\textsuperscript{$\ddagger$} (Wilcoxon signed-rank test, $p < .05$). Inter-annotator agreement (IAA) was calculated using Krippendorff’s alpha.
	\label{tab:human_study_eli5}}
\end{table*}

\begin{table*}[t]
	\small
	\centering
	\begin{tabular}{llrrrrrr}
		\toprule
		&  &
		\multicolumn{2}{c}{\textbf{Steering Success}} &
		\multicolumn{2}{c}{\textbf{Plausibility}} &
		\multicolumn{2}{c}{\textbf{Helpfulness}} \\
		\cmidrule(lr){3-4}
		\cmidrule(lr){5-6}
		\cmidrule(lr){7-8}
		\textbf{Prompt Components} & \textbf{Steering Vectors} & \textbf{Mean{$\;\pm$std}} & \textbf{IAA} & \textbf{Mean{$\;\pm$std}} & \textbf{IAA} & \textbf{Mean{$\;\pm$std}} & \textbf{IAA} \\
		\midrule
		group & - & \textsuperscript{$\dagger\!$}3.29$\;\pm1.10$ & 0.413 & \textsuperscript{$\dagger\!$}4.57$\;\pm0.89$ & 0.387 & 4.01$\;\pm1.10$ & 0.489 \\
		group & \citet{konen2024steeringstyle} & \textsuperscript{$\ddagger\!$}3.53$\;\pm1.38$ & 0.521 & \textsuperscript{$\ddagger\!$}2.85$\;\pm1.48$ & 0.460 & \textsuperscript{$\ddagger\!$}2.71$\;\pm1.32$ & 0.444 \\
		\textbf{group + attributes (5)} & \textbf{Our approach (5)} & \textsuperscript{$\dagger \ddagger\!$}4.35$\;\pm0.84$ & 0.269 & \textsuperscript{$\dagger \ddagger\!$}4.12$\;\pm1.00$ & 0.257 & \textsuperscript{$\ddagger\!$}3.90$\;\pm1.03$ & 0.312 \\
		\bottomrule
	\end{tabular}
	\caption{Manual evaluation results based on 25 randomly sampled science questions: We evaluate \textit{steering success} in terms of how well the explanations are tailored to the target group, the \textit{plausibility} of the explanations and, how \textit{helpful} the explanations are to understand the topic generated by Qwen 7B for the two baselines and our approach. Significant difference to the prompting baseline are marked with \textsuperscript{$\dagger$} and significant difference to  \citet{konen2024steeringstyle} are marked with~\textsuperscript{$\ddagger$} (Wilcoxon signed-rank test, $p < .05$). Inter-annotator agreement (IAA) was calculated using Krippendorff’s alpha.
		\label{tab:human_study_rooein}}
\end{table*}

\subsection{Explanation Samples}
\label{appendix:samples}
Table \ref{tab:samples-1-prompting} - \ref{tab:samples-2-approach} show the generated explanations for two questions generated by Qwen 7B for the two baselines and our approach. Upon manual analysis, we find that the explanations of the baseline based on \citet{konen2024steeringstyle} are somewhat tailored to the target group but do not answer the question anymore (e.g., code-based explanation in Table \ref{tab:samples-1-konen}). This might explain the low helpfulness scores. In contrast, the explanations of the prompting baseline perfectly answer the questions but are barley tailored the target group. As shown in the human study, our approach demonstrates the best steering success by successfully tailoring the explanations to the target group (Table~\ref{tab:samples-1-approach} and \ref{tab:samples-2-approach}). Given that higher steering success does not appear to translate into greater helpfulness, future work should focus on determining when successful tailoring is helpful and when it is not, in order to ensure that explanations are both tailored and helpful. For example, rather simple topics could already be well-understood with generic explanations as demonstrated by the prompting baseline.

\begin{table*}[ht]
	\centering
	\small
	\begin{tabular}{l|p{12.5cm}}
		\toprule
		\textbf{Group} & \textbf{Question: How does DNA work?} \\
		\midrule
		
		\multirow{5}{*}{Biologists}
		& DNA, or deoxyribonucleic acid, is the molecule that carries genetic information in living organisms. It's structured as a double helix, with two strands coiled around each other. Each strand is made of nucleotides, which consist of a sugar (deoxyribose), a phosphate group, and one of four nitrogenous bases: adenine (A), thymine (T), cytosine (C), and guanine (G).
		
		The sequence of these bases along the DNA strand encodes instructions for making proteins, which are essential for life. The base pairs (A-T and C-G) hold the two strands together. When cells divide, enzymes help unzip the DNA, allowing each strand to serve as a template for making a new complementary strand. This process, called replication, ensures genetic information is passed on.
		
		In gene expression, specific sequences of DNA are transcribed into RNA, which then guides the synthesis of proteins through translation. This process allows cells to read and use the genetic information stored in DNA to perform various functions.\\
		\midrule
		
		\multirow{5}{*}{Game developers}
		& DNA, or deoxyribonucleic acid, is the molecule that carries genetic information in living organisms. It's structured as a double helix, with two strands coiled around each other. Each strand is made of nucleotides, which contain a sugar, a phosphate group, and one of four nitrogenous bases: adenine (A), thymine (T), cytosine (C), and guanine (G).
		
		The sequence of these bases along the DNA strand encodes instructions for making proteins, which are essential for life. The base pairs (A-T and C-G) hold the two strands together. When cells divide, enzymes read the DNA sequence, transcribe it into RNA, and then translate the RNA into proteins.
		
		In games, you might use DNA analogies for character customization, evolution systems, or genetic traits that affect gameplay mechanics.\\
		\midrule
		
		\multirow{5}{*}{Philosophers}
		& DNA, or deoxyribonucleic acid, is a molecule that encodes the genetic instructions used in the development and functioning of all known living organisms. It works through a process of replication, transcription, and translation:
		
		1. **Replication**: During cell division, DNA replicates itself, ensuring each new cell receives a complete copy.
		
		2. **Transcription**: In the nucleus, enzymes transcribe DNA into messenger RNA (mRNA), which carries genetic information from DNA to the ribosomes.
		
		3. **Translation**: Ribosomes read the mRNA and synthesize proteins according to the genetic code, using transfer RNA (tRNA) to match codons with amino acids.
		
		This process allows for the transmission of genetic information and the synthesis of proteins essential for life. \\
		\midrule
		
	\end{tabular}
	\caption{Explanation samples of the \emph{prompting baseline} generated by Qwen 7B for the question ``How does DNA work?'' tailored to three different target groups $g \in \{biologists, game developers, philosophers\}$. The prompting 
		baseline uses the target group $g$ in the prompt.\label{tab:samples-1-prompting}}
\end{table*}

\begin{table*}[ht]
	\centering
	\small
	\begin{tabular}{l|p{12.5cm}}
		\toprule
		\textbf{Group} & \textbf{Question: How does DNA work?} \\
		\midrule
		
		\multirow{5}{*}{Biologists}
		&  DNA, or deoxyribonucleic acid, is a double-stranded molecule that contains the genetic instructions used in the development and functioning of all living organisms. It is organized into units called genes, which provide instructions for specific proteins. DNA consists of four nucleotide bases: adenine (A), cytosine (C), guanine (G), and thymine (T). These bases pair up to form base pairs, with A pairing with T and C pairing with G.
		
		DNA replication occurs during cell division, where each strand serves as a template for the synthesis of a complementary strand. This process is crucial for inheritance and is highly conserved across different species. Additionally, DNA undergoes transcription, where the genetic information is transcribed into RNA molecules, and translation, where these RNA molecules are translated into proteins. These processes are essential for various cellular functions and are regulated by complex mechanisms.\\
		\midrule
		
		\multirow{5}{*}{Game developers}
		& \begin{minipage}{12.5cm}
			Sure! Here’s a simple way to set up your DNA system:
			
			1. Structure: Use a basic structure like this:
			
			\begin{verbatim}
				public class DNA
				public class DNA {
					private List<Vector2> points;
				}
			\end{verbatim}
			
			Make sure to include methods for adding, removing, or updating points based on player input or logic.
			
			2. Example:
			
			\begin{verbatim}
				public class DNA {
					private List<Vector2> points;
					
					public DNA() {
						points = new List<Vector2();
					}
					
					public void AddPoint(Vector2 point) {
						points.Add(point);
					}
					
					public void RemovePoint(int index) {
						if (index >= 0 && index < points.Count) {
							points.RemoveAt(index);
						}
					}
				}
			\end{verbatim}
			
			3. Don’t forget to clear the list when you want to reset or reuse the DNA:
			
			\begin{verbatim}
				public void Clear() {
					points.Clear();
				}
			\end{verbatim}
		This will give you a flexible DNA system where you can easily manage your points for things like player movement or animations.
			
		\end{minipage}\\
		\midrule
		
		\multirow{5}{*}{Philosophers}
		&  DNA, as a philosophical inquiry, is not reducible to mere biological explanation but invites us to explore its ontological and epistemological dimensions. DNA, as a code or a narrative, suggests a narrative of being, a story of what we are and what we are not. It is a system of signs, a language that gestures toward the possibility of meaning, where the 'is' of identity is not exhausted in the mere assertion of being but in the articulation of a narrative that is both constitutive and constituting. In this sense, DNA is not a brute fact but a threshold, a boundary between the self and the world, a limit that marks the limits of thought. To know DNA is to engage with the question of what it is to be a subject, to be an agent, to be a knower, and to be a narrative that seeks to articulate itself in the face of the mystery of being.
		\\
		\midrule
		
	\end{tabular}
	\caption{Explanation samples of the \emph{activation-based steering baseline} \cite{konen2024steeringstyle} generated by Qwen 7B for the question ``How does DNA work?'' tailored to three different target groups $g \in~\{biologists, game developers, philosophers\}$. The baseline uses the target group $g$ in the prompt and the activation-based steering vectors based on the average of the activation vectors extracted for the 500 Stack Exchange answers of $g$.\label{tab:samples-1-konen}}
\end{table*}

\begin{table*}[ht]
	\centering
	\small
	\begin{tabular}{l|p{12.5cm}}
		\toprule
		\textbf{Group} & \textbf{Question: How does DNA work?} \\
		\midrule
		
		\multirow{5}{*}{Biologists}
		&  DNA, or deoxyribonucleic acid, is the molecule that carries the genetic instructions used in the development, functioning, reproduction, and evolution of all known living organisms. At its core, DNA is composed of four types of nucleotides—adenine (A), thymine (T), cytosine (C), and guanine (G)—which pair up to form the rungs of the double helix structure. The sequence of these nucleotides along the DNA strands encodes the information needed for proteins to be synthesized, which is crucial for the proper functioning of cells.
		
		In molecular biology, we often discuss how DNA replicates itself during cell division, ensuring that genetic information is passed on accurately. This process involves enzymes like DNA polymerase, which synthesizes new strands by adding nucleotides complementary to the template strand. Mutations, which can occur spontaneously or due to external factors like radiation or chemicals, introduce variations in the DNA sequence, which can lead to changes in traits and sometimes contribute to evolution.
		
		From an evolutionary perspective, DNA plays a central role in how species adapt to their environments over time. Genetic variations arising from mutations provide the raw material for natural selection, allowing populations to evolve in response to selective pressures. Understanding these processes helps us delve into how different organisms function at the genetic level and how they respond to various environmental challenges.\\
		\midrule
		
		\multirow{5}{*}{Game developers}
		& From a game developer's perspective, you might want to think of DNA as a blueprint that defines the structure and behavior of your game. Just like DNA codes for the traits of an organism, your game's code defines its behavior and features. For instance, you might have functions that initialize game objects, handle user input, or manage game states. These functions are like the genes that determine how your game functions at different stages. You could also consider using comments to explain complex parts of your code, much like DNA sequences are annotated to provide context. This helps in making the code more understandable and maintainable, enhancing the overall gaming experience.\\
		\midrule
		
		\multirow{5}{*}{Philosophers}
		& From a philosophical perspective, DNA can be seen as a fundamental aspect of life and existence, embodying the essence of biological identity and heredity. Philosophers might explore how DNA shapes our understanding of identity over time, from generation to generation, and how it influences our sense of self and continuity. They might also delve into the ethical implications of genetic modification and the role of DNA in defining what it means to be human in a technological age. This exploration can lead to deeper questions about the nature of life itself and the role of biology in shaping human experience and society. \\
		\midrule
		
	\end{tabular}
	\caption{Explanation samples of \emph{our approach} generated by Qwen 7B for the question ``How does DNA work?'' tailored to three different target groups $g \in \{biologists, game developers, philosophers\}$. Our approach uses the five most important attributes identified for $g$ in the prompt and the attribute-based steering vector according to Equation~\ref{eq:steeringvector}.\label{tab:samples-1-approach}}
\end{table*}

\begin{table*}[ht]
	\centering
	\small
	\begin{tabular}{l|p{12.5cm}}
		\toprule
		\textbf{Group} & \textbf{Question: What is the role of enzymes in digestion?} \\
		\midrule
		
		\multirow{5}{*}{Biologists}
		& Enzymes play a crucial role in digestion by catalyzing specific biochemical reactions that break down complex food molecules into simpler forms that can be absorbed by the body. Each enzyme is highly specialized, targeting particular types of molecules like proteins, carbohydrates, or fats.
		
		For example, pepsin breaks down proteins in the stomach, while amylase starts breaking down carbohydrates in the mouth and small intestine. Lipases break down fats into fatty acids and glycerol. These enzymes lower the activation energy required for these reactions to occur, making them much faster and more efficient.
		
		By facilitating these breakdown processes, enzymes ensure that nutrients are available for absorption and utilization by cells throughout the body, supporting overall health and metabolic functions.
		\\
		\midrule
		
		\multirow{5}{*}{Game developers}
		& Enzymes in digestion play a crucial role by breaking down complex food molecules into simpler forms that the body can absorb. For example, amylase breaks down starch into sugars, proteases break proteins into amino acids, and lipases break fats into fatty acids and glycerol. These enzymes work in specific environments (like the stomach or small intestine) with optimal pH and temperature conditions. They speed up chemical reactions without being consumed themselves, making digestion efficient and effective. Understanding enzyme functions helps in designing realistic food interactions and health effects in games.\\
		\midrule
		
		\multirow{5}{*}{Philosophers}
		& Enzymes play a crucial role in digestion by catalyzing biochemical reactions that break down complex food molecules into simpler forms, facilitating absorption. Each enzyme is specific to its substrate, ensuring efficient and targeted breakdown. For instance, amylase breaks down starch into sugars, proteases digest proteins into amino acids, and lipases convert fats into fatty acids and glycerol. This process not only aids in the extraction of nutrients but also reduces the energy required for the body to perform these chemical transformations on its own. Thus, enzymes are indispensable in the digestive process, optimizing nutrient availability and supporting overall metabolic functions. \\
		\midrule
		
	\end{tabular}
	\caption{Explanation samples of the \emph{prompting baseline} generated by Qwen 7B for the question ``What is the role of enzymes in digestion?'' tailored to three different target groups $g \in \{biologists, game developers, philosophers\}$. The prompting 
		baseline uses the target group $g$ in the prompt.\label{tab:samples-2-prompting}}
\end{table*}

\begin{table*}[ht]
	\centering
	\small
	\begin{tabular}{l|p{12.5cm}}
		\toprule
		\textbf{Group} & \textbf{Question: What is the role of enzymes in digestion?} \\
		\midrule
		
		\multirow{5}{*}{Biologists}
		& Enzymes play crucial roles in digestion by catalyzing the breakdown of complex molecules into simpler forms, which can be absorbed by the body. These enzymes are produced by various organs and cells in the gastrointestinal tract, including the salivary glands, pancreas, and small intestine. They act as biological catalysts that increase the rate of chemical reactions without being consumed or altered during the process.
		
		1. Amylases: Break down carbohydrates into simple sugars.
		2. Lipases: Break down lipids (fats) into fatty acids and glycerol.
		3. Proteases: Break down proteins into amino acids.
		4. Nucleases: Break down nucleic acids into nucleotides.
		
		These enzymes help in the breakdown of macromolecules into smaller molecules that can be further metabolized for energy or used as building blocks for cellular functions. \\
		\midrule
		
		\multirow{5}{*}{Game developers}
		& Sure! Enzymes are super useful for handling your game's input or processing logic. Think of them like little functions that you can reuse to handle specific tasks. For example, if you want to add a button for jumping, you can create an enzyme (or function) like `Jump()` and call it whenever the player clicks the jump button. This way, you can easily manage inputs and keep your code clean and organized. Make sure to pass any necessary parameters to your enzyme, so it knows what to do based on the input.\\
		\midrule
		
		\multirow{5}{*}{Philosophers}
		& Enzymes, as philosophical inquiries might suggest, are not mere tools but rather the locus of a profound inquiry into the nature of meaning and the structure of thought. Enzymes, like Kantian noumena, are not objects of thought but are rather the conditions of thought itself, for they are the necessary conditions for the intelligibility of experience, just as the conditions of thought are necessary for the intelligibility of experience. In the domain of philosophy, to ask about the role of enzymes in digestion is to inquire into the conditions of our understanding of the world, for it is through the act of thinking, through the act of enacting thought, that we encounter the world as a system of meanings, as a web of relations, as a network of thought. Thus, to explore the role of enzymes in digestion is to explore the limits of thought, the boundaries of the self, and the horizon of the human condition. Enzymes, in this sense, are not merely cognitive or epistemological, but are the very ground of our being, the very soil from which thought springs.\\
		\midrule
		
	\end{tabular}
	\caption{Explanation samples of the \emph{activation-based steering baseline} \cite{konen2024steeringstyle} generated by Qwen 7B for the question ``What is the role of enzymes in digestion?'' tailored to three different target groups $g \in~\{biologists, game developers, philosophers\}$. The baseline uses the target group $g$ in the prompt and the activation-based steering vectors based on the average of the activation vectors extracted for the 500 Stack Exchange answers of $g$.\label{tab:samples-2-konen}}
\end{table*}

\begin{table*}[ht]
	\centering
	\small
	\begin{tabular}{l|p{12.5cm}}
		\toprule
		\textbf{Group} & \textbf{Question: What is the role of enzymes in digestion?} \\
		\midrule
		
		\multirow{5}{*}{Biologists}
		&  Enzymes play a crucial role in digestion by catalyzing biochemical reactions that break down food into components that can be absorbed by the body. For instance, amylase enzymes break down carbohydrates into sugars, lipases break down fats into fatty acids and glycerol, and proteases break down proteins into amino acids. These enzymes work optimally within specific pH ranges; for example, pepsin, a protease, functions best in the acidic environment of the stomach, while pancreatic lipase works efficiently in the alkaline environment of the small intestine. The efficiency of these enzymes is essential for breaking down complex molecules into simpler forms that can be easily absorbed, which is vital for maintaining proper nutrition and overall health. Understanding enzyme function also helps in diagnosing and treating various conditions, such as lactose intolerance or malabsorption syndromes, where specific enzymes might not be functioning properly.\\
		\midrule
		
		\multirow{5}{*}{Game developers}
		& From a game developer's perspective, let's think about how we can explain the role of enzymes in digestion in a way that enhances your understanding and can be applied within the context of a game or simulation. Enzymes in digestion can be likened to functions in programming—each has a specific role that, when executed correctly, ensures smooth operation. Just as functions handle specific tasks in a program, enzymes handle specific tasks in the body, breaking down food into nutrients that the body can use. In a game, you might simulate this process by having different functions or methods that handle various aspects of digestion, such as breaking down food into simpler components or managing the absorption of nutrients. This can help players understand the mechanics of the game more deeply, just as understanding enzymes helps players understand the biological processes in the body.\\
		\midrule
		
		\multirow{5}{*}{Philosophers}
		& From a philosophical perspective, enzymes in digestion can be seen as agents of transformation, much like how philosophers transform ideas and concepts. Just as enzymes catalyze chemical reactions in the body, philosophers catalyze change through their analysis and synthesis of thoughts and theories. Enzymes, like philosophical inquiry, are essential for breaking down complex structures into simpler components, allowing for better understanding and assimilation. Similarly, philosophers break down complex problems into manageable parts to explore deeper truths. The role of enzymes in digestion, then, mirrors the role of philosophical methods in understanding the world. \\
		\midrule
		
	\end{tabular}
	\caption{Explanation samples of \emph{our approach} generated by Qwen 7B for the question ``What is the role of enzymes in digestion?'' tailored to three different target groups $g \in \{biologists, game developers, philosophers\}$. Our approach uses the five most important attributes identified for $g$ in the prompt and the attribute-based steering vector according to Equation~\ref{eq:steeringvector}.\label{tab:samples-2-approach}}
\end{table*}

\end{document}